\documentclass[10pt,twocolumn,letterpaper]{article}

\usepackage[pagenumbers]{iccv} 

\usepackage{colortbl}
\usepackage{multirow}
\usepackage{makecell}
\usepackage{arydshln}
\usepackage[normalem]{ulem}

\definecolor{iccvblue}{rgb}{0.21,0.49,0.74}
\usepackage[pagebackref,breaklinks,colorlinks,allcolors=iccvblue]{hyperref}
\definecolor{myred}{RGB}{249, 221, 223} 
\definecolor{myblue}{RGB}{221, 222, 237} 

\def\paperID{390} 
\def\confName{ICCV}
\def\confYear{2025}

\title{Revisiting Image Fusion for Multi-Illuminant White-Balance Correction} 

\author{David Serrano-Lozano$^{1,2}$ \quad Aditya Arora$^{3,4}$ \quad Luis Herranz$^{5}$ \quad Konstantinos G. Derpanis$^{3,4}$\\Michael S. Brown$^{3}$ \quad Javier Vazquez-Corral$^{1,2}$\\
{\normalsize $^1$Computer Vision Center} \quad
{\normalsize $^2$Universitat Autònoma de Barcelona} \quad
{\normalsize $^3$York University}\\
{\normalsize $^4$Vector Institute \quad $^5$Universidad Autónoma de Madrid}\\
{\normalsize \texttt{\{dserrano, javier.vazquez\}@cvc.uab.cat \quad luis.herranz@uam.es}}\\
{\normalsize \texttt{\{adityac8, kosta, mbrown\}@yorku.ca}}
}

\begin{document}
\twocolumn[{%
\renewcommand\twocolumn[1][]{#1}%
\maketitle
\centering
\captionsetup{type=figure}
    \vspace{-6pt}
    \begin{subfigure}{0.16\linewidth}
        \includegraphics[width=\linewidth]{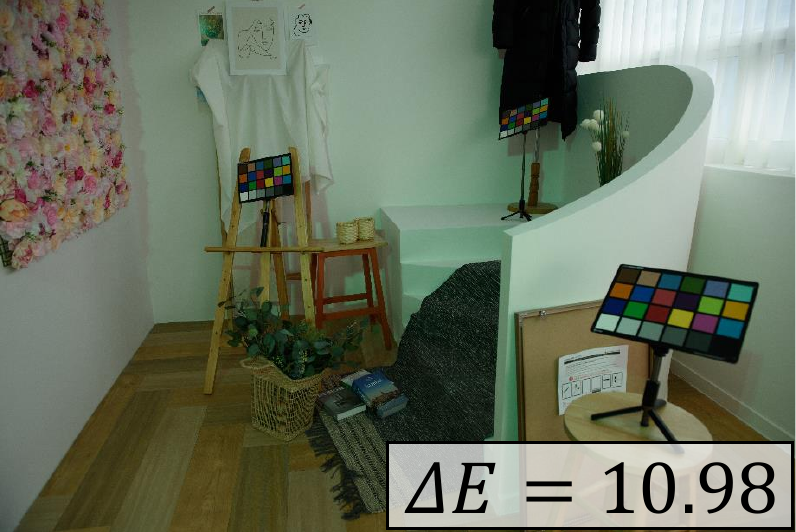}
        \caption{Daylight (5500K)}
    \end{subfigure}
    \begin{subfigure}{0.16\linewidth}
        \includegraphics[width=\linewidth]{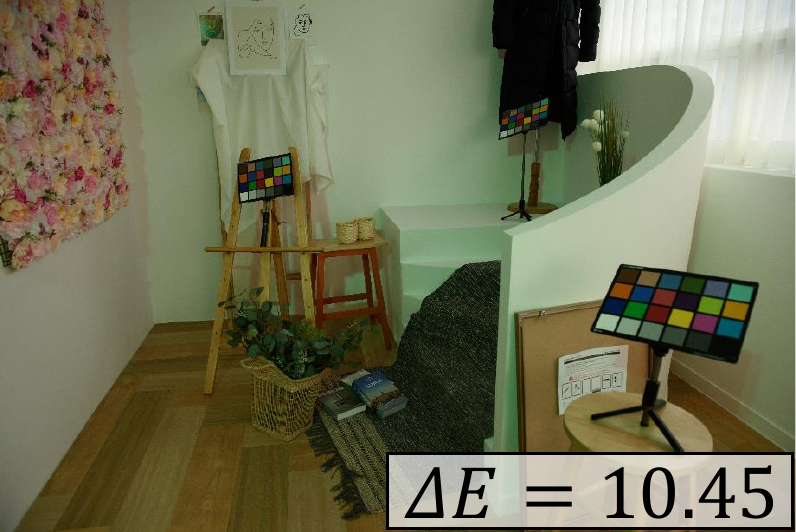}
        \caption{Cloudy (6500K)}
    \end{subfigure}
    \begin{subfigure}{0.16\linewidth}
        \includegraphics[width=\linewidth]{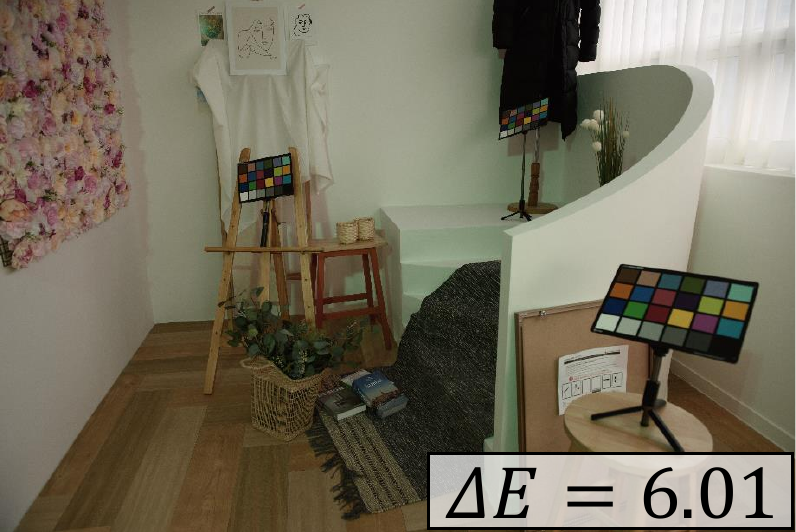}
        \caption{DeepWB~\cite{afifi2020deep}}
    \end{subfigure}
    \begin{subfigure}{0.16\linewidth}
        \includegraphics[width=\linewidth]{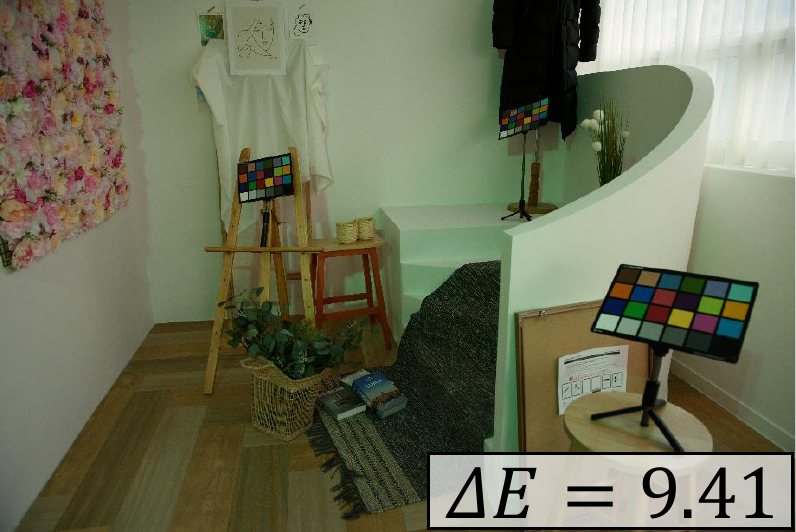}
        \caption{MixedWB~\cite{afifi2022}}
    \end{subfigure}
    \begin{subfigure}{0.16\linewidth}
        \includegraphics[width=\linewidth]{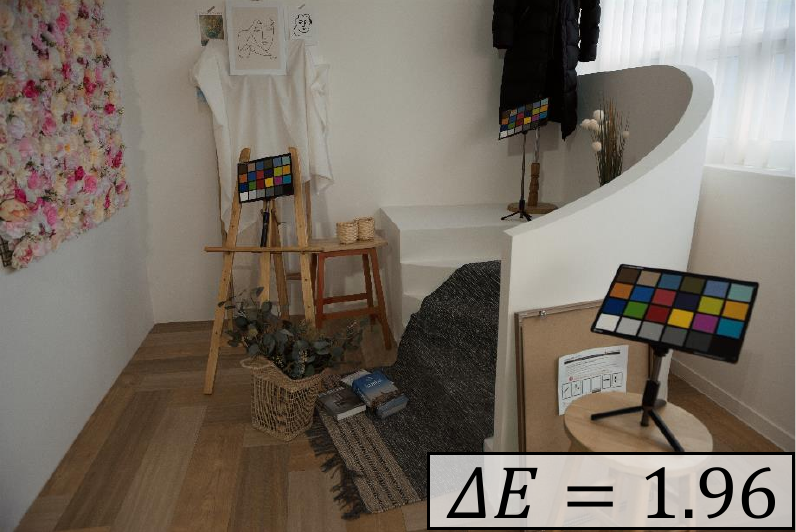}
        \caption{Ours}
    \end{subfigure}
    \begin{subfigure}{0.16\linewidth}
        \includegraphics[width=\linewidth]{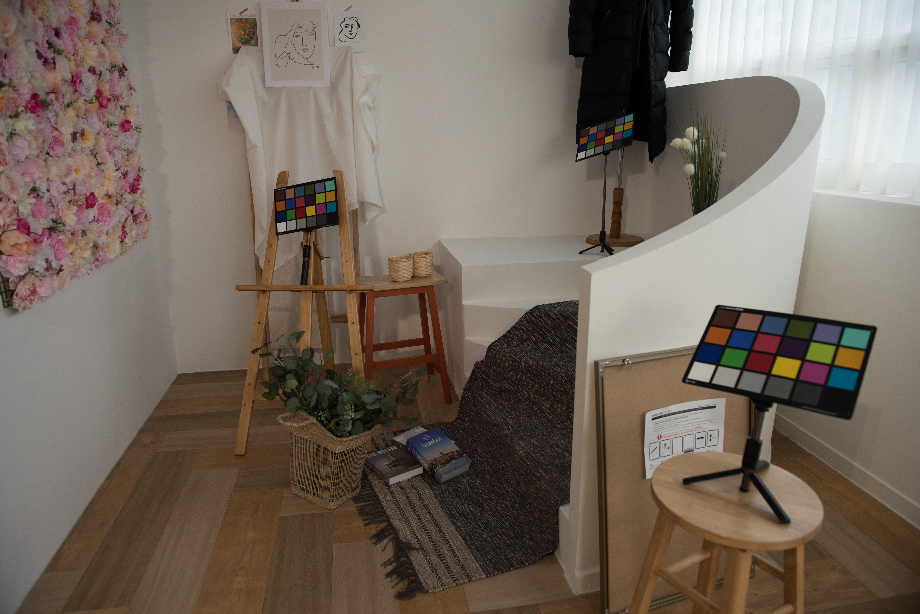}
        \caption{White-balanced}
    \end{subfigure}
\vspace{-1mm}
\caption{Camera white balance (WB) presets often struggle to correct colors in scenes with multiple illuminants, leading to undesirable color tints in the final sRGB image, as demonstrated in (a) and (b). Recent WB methods, such as DeepWB~\cite{afifi2020deep}, also fail in mixed-illumination scenarios, as shown in (c). MixedWB~\cite{afifi2022} addressed this issue by estimating a pixel-wise weight map to linearly blend multiple WB settings. However, their method can fail to achieve proper color correction in complex scenes, as seen in the white wall of (d). In contrast, we present a new efficient transformer-based method to blend five WB settings end-to-end. Our approach successfully handles white balance in scenes with several illuminants, as shown in (e). Finally, we show the white-balanced image in (f). Each image's $\Delta$E2000 values are displayed in the bottom-right corner (lower values indicate higher performance).}
\vspace{10mm}
\label{fig:teaser}
}]

\begin{abstract}
White balance (WB) correction in scenes with multiple illuminants remains a persistent challenge in computer vision. Recent methods explored fusion-based approaches, where a neural network linearly blends multiple sRGB versions of an input image, each processed with predefined WB presets. However, we demonstrate that these methods are suboptimal for common multi-illuminant scenarios. Additionally, existing fusion-based methods rely on sRGB WB datasets lacking dedicated multi-illuminant images, limiting both training and evaluation. To address these challenges, we introduce two key contributions. First, we propose an efficient transformer-based model that effectively captures spatial dependencies across sRGB WB presets, substantially improving upon linear fusion techniques. Second, we introduce a large-scale multi-illuminant dataset comprising over 16,000 sRGB images rendered with five different WB settings, along with WB-corrected images. Our method achieves up to 100\% improvement over existing techniques on our new multi-illuminant image fusion dataset. 
\end{abstract}

\vspace{-14pt}
\section{Introduction}\label{sec:intro}

Auto-white-balance (AWB) correction is a crucial step applied onboard cameras to remove the color cast caused by the light conditions in the imaged scene.  Proper white-balance correction is needed to produce visually pleasing images and to assist downstream computer vision tasks~\cite{afifi2019color}.

Conventional AWB techniques \cite{BUCHSBAUM19801, DBLP:journals/tip/GijsenijGW11, vazquez2011color, DBLP:conf/cvpr/HuWL17, DBLP:conf/cvpr/BarronT17} assume that a \emph{single} light source illuminates a scene. Typical AWB methods operate on the RAW sensor image to first estimate the color cast in the image in the form of a sensor-specific $R$,$G$,$B$ vector. Subsequently, the illumination's color cast is removed by scaling each color channel of the RAW image such that the estimated illumination becomes achromatic (i.e., $R$=$G$=$B$). The white-balanced RAW image is then rendered through the camera's image signal processor (ISP) to render the final image in a standard RGB (sRGB) space.

While the single-illuminant assumption works reasonably well for many scenes, it fails on common scenes with multiple illuminants, such as indoor scenes with illumination from indoor and outdoor lighting (e.g., via a window).
Prior work examined ways to correct such multiple illuminations~\cite{sidorov2019conditional,vrvsnak2022illuminant,vrvsnak2022autoencoder}. These methods typically follow a two-step process that first estimates the constituent scene illuminants, followed by spatially varying WB correction in the RAW sensor image based on the estimates.  The need for both estimation and correction makes this a challenging problem.  Another bottleneck for multi-illuminant learning methods is the difficulty in preparing per-pixel annotated RAW image datasets for training and testing.

Afifi et al.~\cite{afifi2022} introduced MixedWB, a multi-illuminant WB correction approach that leverages pre-defined WB values for common scene illuminations (i.e., the camera settings for {\it manual WB mode} rather than AWB). The sensor's $R$,$G$,$B$ response to several common scene illuminations can be accessed directly via camera API calls or image metadata. Given a RAW input image, MixedWB first renders it to sRGB using five WB settings --- {\it tungsten}, {\it fluorescent}, {\it daylight}, {\it cloudy}, and {\it shade}. A neural network then predicts per-pixel blending weights to linearly merge these five images into a final sRGB output. This formulation transforms the inherently ill-posed problem of per-pixel illuminant estimation into a more straightforward task: learning a combination of a fixed set of images. 

MixedWB, however, has two main limitations. First, while linear blending is effective in the RAW linear color space, MixedWB applies it in the non-linear sRGB space. In this paper, we show that this approach is suboptimal because ground-truth correct pixels in sRGB often fall outside the convex hull defined by the preset pixels. To address this limitation, we propose a transformer-based~\cite{vaswani2017attention} fusion strategy that learns a more flexible and accurate WB blending function.

Second, MixedWB is trained on the RenderedWB dataset~\cite{afifi2019color}, which contains images rendered with distinct sRGB WB presets but lacks true multi-illuminant scenes, limiting its effectiveness for real-world WB correction. The only available true multi-illuminant sRGB dataset is generated from 3D-rendered scenes~\cite{afifi2022}, introducing a domain shift from real-world images. To address this limitation, we introduce a new multi-illuminant dataset containing 16,284 sRGB images, each rendered with five WB presets and accompanied by a corresponding ground-truth WB-corrected image.

Figure~\ref{fig:teaser} illustrates the benefits of our approach using a multi-illuminant scene rendered with the two most suitable WB presets ({\it Daylight} and {\it Cloudy}). We compare the results of an sRGB WB method (DeepWB~\cite{afifi2020deep}), MixedWB~\cite{afifi2022}, our transformer-based approach, and the ground-truth white-balanced image. The results show that all WB methods struggle to remove the greenish color cast, particularly noticeable on the white wall, except for our transformer-based method, which achieves the most accurate correction.


\vspace{-4mm}
\paragraph{Contribution.}~We revisit fusion-based multi-illumination WB correction and introduce two key contributions. First, we propose a transformer-based method that performs WB in multi-illuminant scenes more accurately and efficiently by blending five distinct WB presets. Second, we present a new multi-illuminant white balance dataset in the sRGB domain, created by repurposing the RAW multi-illuminant dataset from Kim et al.~\cite{kim2021large}. This dataset comprises 2,714 scenes, each rendered into five WB sRGB images with a corresponding ground-truth white-balanced image. Our method significantly outperforms previous approaches across multiple datasets, including our new multi-illuminant dataset, Cube+~\cite{banic2017unsupervised}, and the Synthetic test set~\cite{afifi2022}.

\begin{figure*}[t!]
\begin{center}
    \begin{minipage}{0.60\linewidth} 
        \centering
        \begin{subfigure}{0.32\linewidth}
            \includegraphics[width=\linewidth]{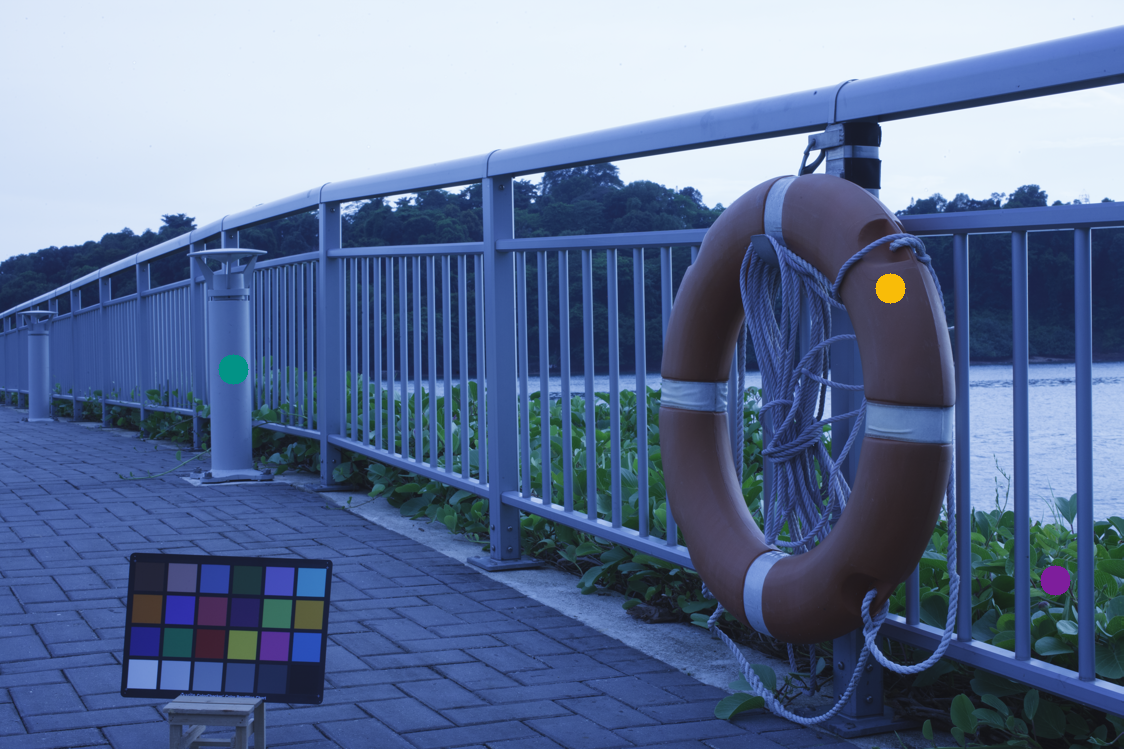}    
            \caption{Tungsten}
        \end{subfigure}
        \begin{subfigure}{0.32\linewidth}
            \includegraphics[width=\linewidth]{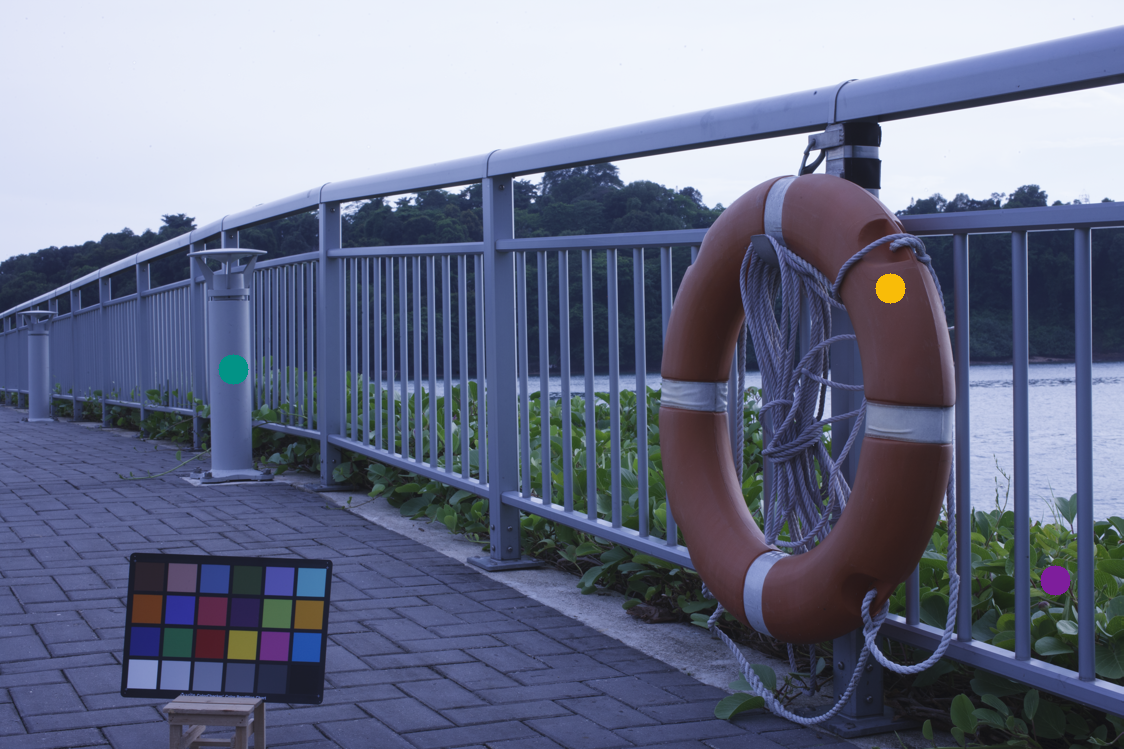}    
            \caption{Fluorescent}
        \end{subfigure}
        \begin{subfigure}{0.32\linewidth}
            \includegraphics[width=\linewidth]{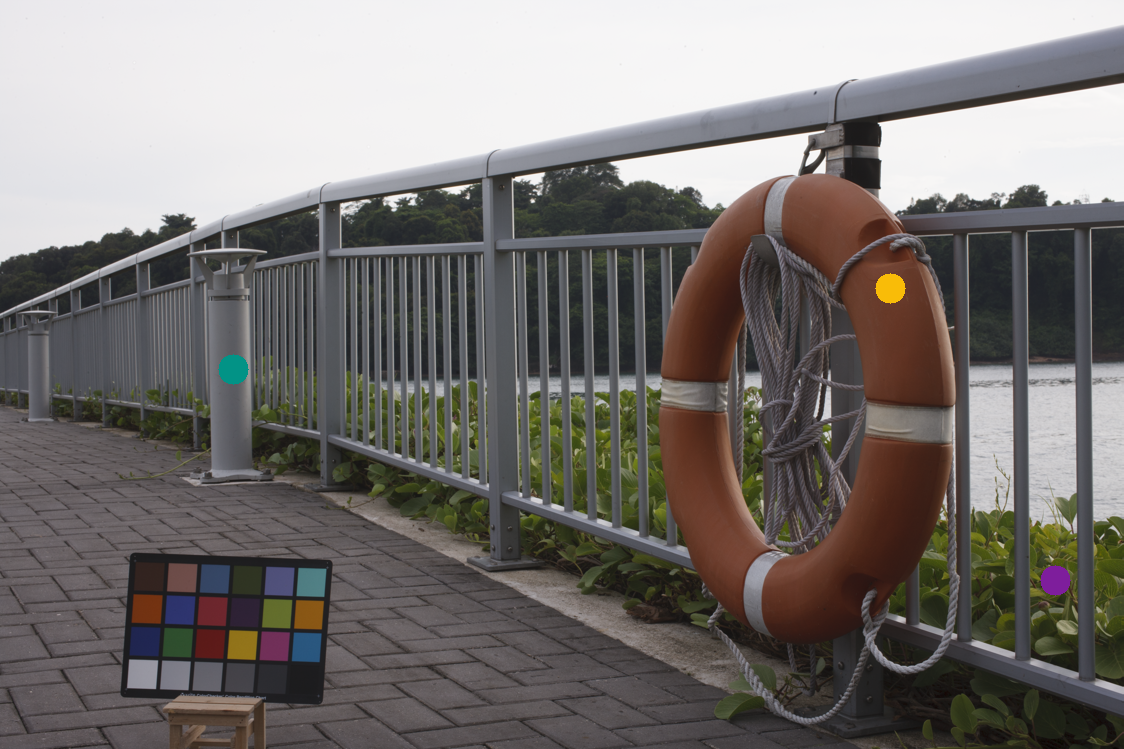}    
            \caption{Daylight}
        \end{subfigure}
        
        \begin{subfigure}{0.32\linewidth}
            \includegraphics[width=\linewidth]{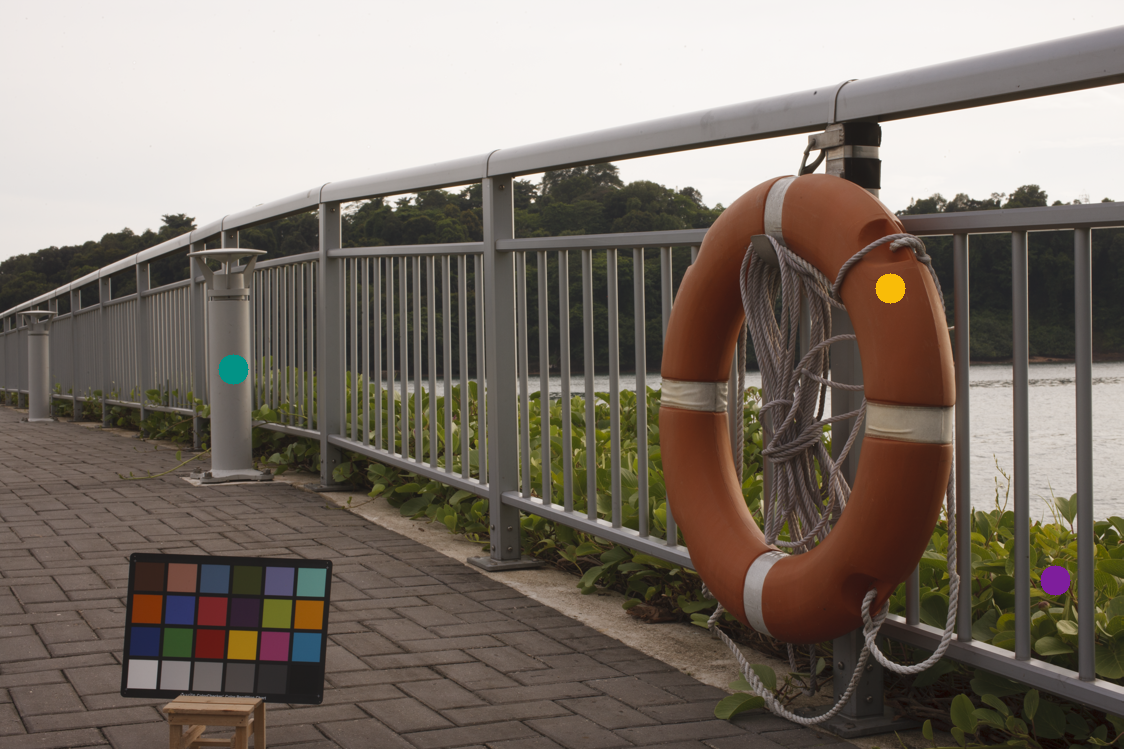}    
            \caption{Cloudy}
        \end{subfigure}
        \begin{subfigure}{0.32\linewidth}
            \includegraphics[width=\linewidth]{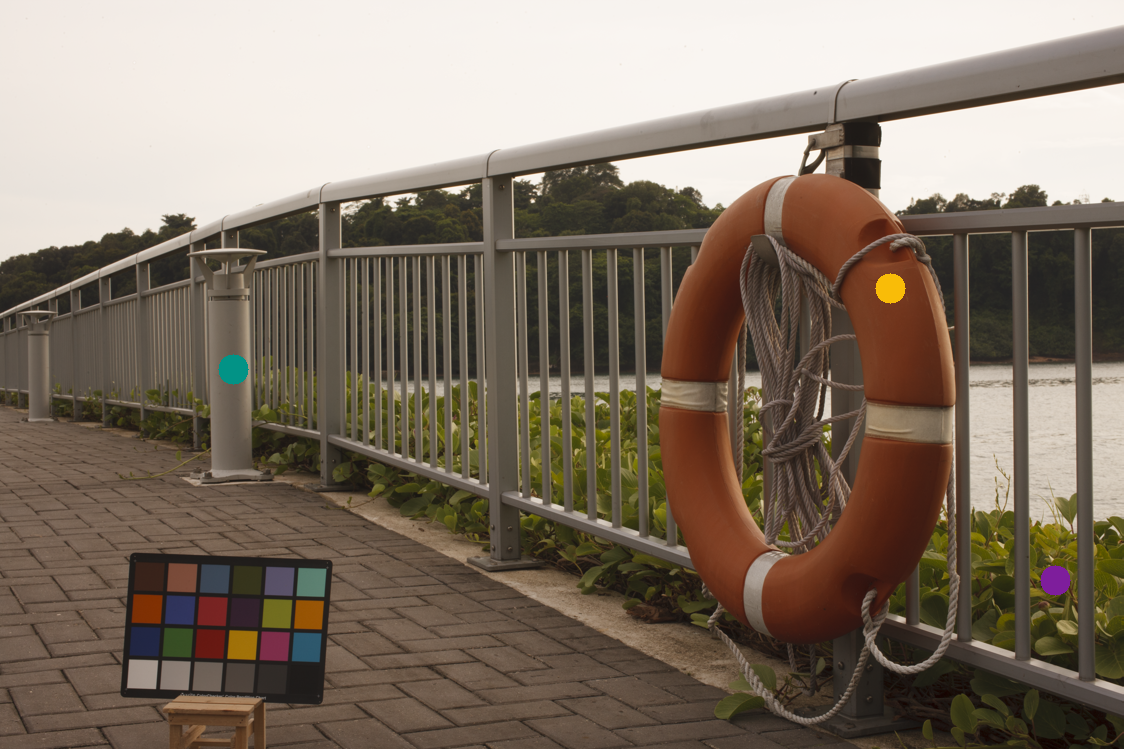}    
            \caption{Shade}
        \end{subfigure}
        \begin{subfigure}{0.32\linewidth}
            \includegraphics[width=\linewidth]{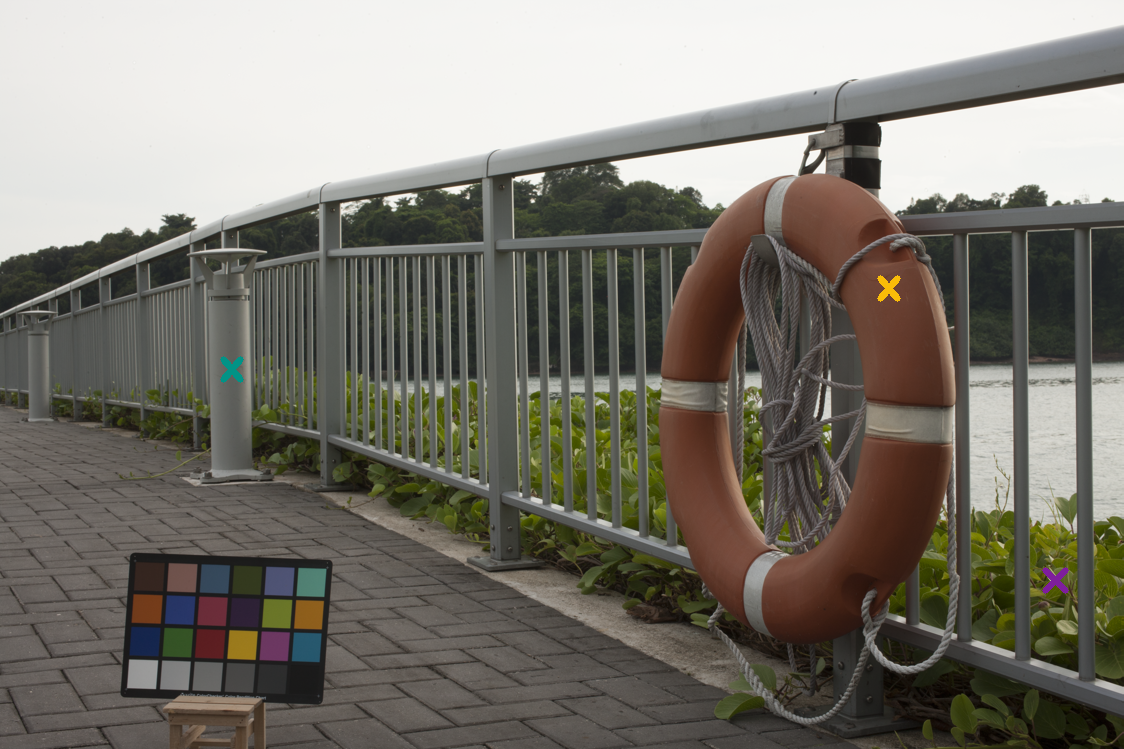}    
            \caption{Ground truth}
        \end{subfigure}
    \end{minipage}
    \begin{minipage}{0.32\linewidth} 
        \centering
        \begin{subfigure}{\linewidth}
            \includegraphics[width=0.9\linewidth]{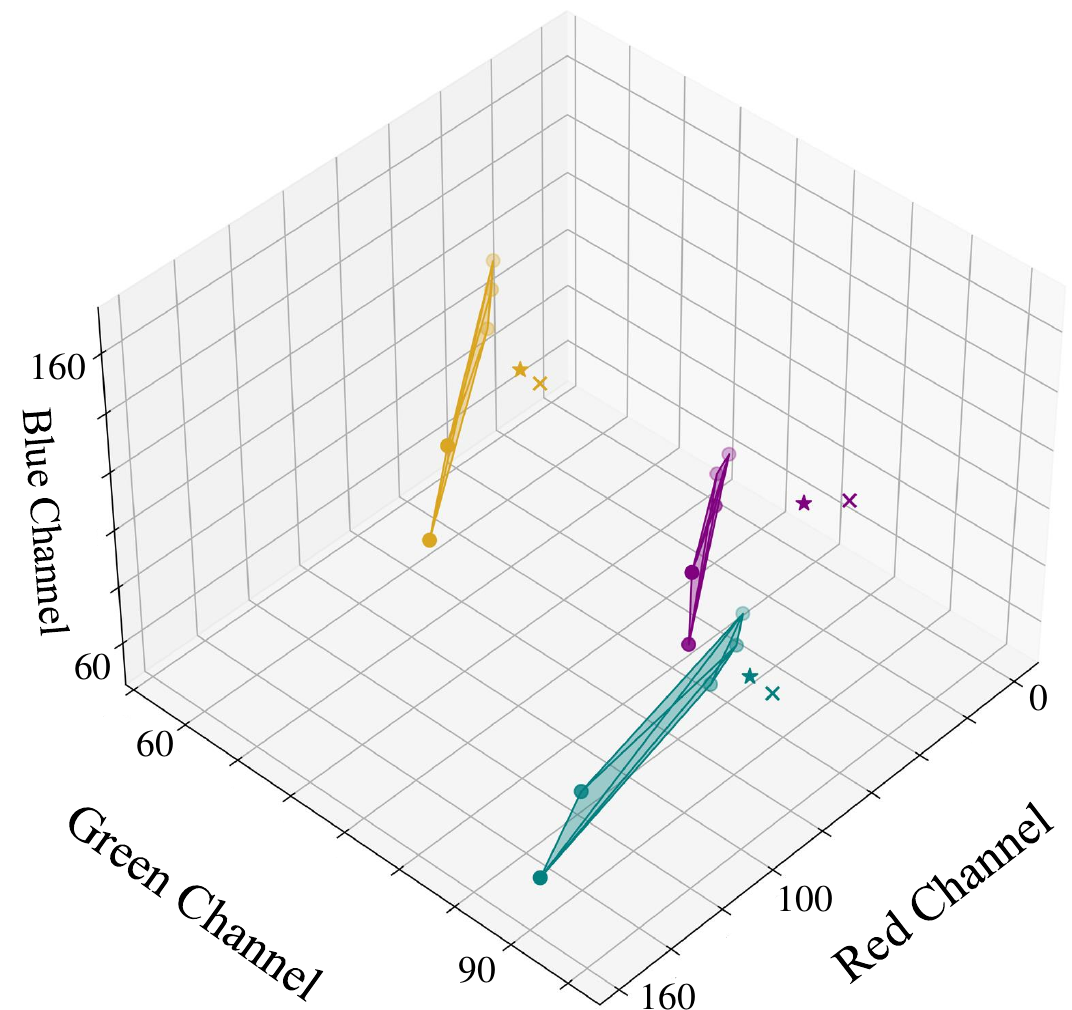}    
            \caption{sRGB cube}
        \end{subfigure}
    \end{minipage}
    \vspace{-6pt}
    \caption{\small Example from the RenderedWB~\cite{afifi2019color} dataset. (a)-(e) show the same scene processed with five distinct WB presets, while (f) presents the white-balanced image obtained using the color checker. Three sample points, marked with teal, yellow, and purple dots for the WB presets and crosses for the ground truth image, are selected across all images. (g) visualizes the pixel values in the sRGB space, along with the polytope formed by the WB presets. In here, we also show the results of our method as stars. Note that each axis has a different scale to ease the visualization.}
    \vspace{-22pt}
    \label{fig:convex_hull}
\end{center}
\end{figure*}

\section{Related work}

Our related work is organized into three areas: (i) conventional WB correction, (ii) WB methods for processed sRGB images, and (iii) multi-illuminant WB correction.

\vspace{-4mm}
\paragraph{Conventional white-balance correction.}~Traditional white balance correction techniques aim to estimate the sensor's response to illumination from a single RAW image.  Early statistical methods~\cite{BUCHSBAUM19801,land1971lightness,finlayson2004shades, van2007edge, chakrabarti2008color, gijsenij2011improving, cheng2014illuminant, finlayson2013corrected, qian2019finding} computed simple RAW image statistics as features for illumination estimation. These early approaches were later extended to employ more complex statistical inferences, such as color-by-correlation \cite{finlayson1997color}, Bayesian color constancy \cite{brainard1997bayesian,gehler2008bayesian}, and voting (e.g., \cite{sapiro1999color,vazquez2009color,vazquez2011color}). 

Deep learning methods now comprise the state of the art for illumination estimation (e.g., \cite{afifi2020interactive,bianco2019quasi,hernandez2020multi, lo2021clcc,lou2015color, oh2017approaching, shi2016deep,xu2020end, zini2022cocoa}).  
Since illumination estimation is performed on the sensor-specific RAW image format, these methods require large, annotated datasets of images for each targeted sensor.

One limitation of conventional methods is their reliance on the assumption of a single dominant light source in the scene. When faced with scenes containing multiple light sources, such as indoor environments with artificial and natural lighting, traditional methods may fail to achieve accurate color reproduction.

\vspace{-4mm}
\paragraph{White-balance methods for rendered sRGB images.}~The camera's ISP hardware generally applies white balance as an early processing step, followed by multiple non-linear adjustments to produce the final sRGB output. This makes post-capture WB correction challenging. Recent works explore methods for performing WB correction directly on sRGB images~\cite{afifi2019color, afifi2019tuning, afifi2020interactive}. A notable advancement in this area is the application of deep neural networks for WB correction and adjustment in camera-rendered sRGB images~\cite{afifi2020deep}. SWBNet~\cite{li2023swbnet} introduced a learning-based architecture to improve the consistency across different color temperatures leveraging a contrastive loss, while WBFlow~\cite{li2023wbflow} proposed a DNN-based method that leverages neural flow. These methods demonstrate improved color accuracy. Still, they often depend on post-capture correction techniques or user interaction, adding complexity and additional steps to the image processing workflow.


\vspace{-2mm}
\paragraph{Multiple-illuminant methods.}~Various techniques target multiple-illuminant white-balance correction. Previous methods often rely on pixel-wise illumination estimation strategies \cite{finlayson1995color,hsu2008light,gijsenij2011color,joze2013exemplar,hussain2018color,beigpour2013multi}, many of which assume that the number of light sources in the scene is known. This strong assumption limits their applicability in real-world scenarios. More recent approaches~\cite{kim2024attentive, yue2024robust} leverage deep learning to estimate pixel-wise illuminants in multi-illuminant scenes directly from the RAW images. 

Most closely related to our work are methods that merge multiple sRGB images. MixedWB~\cite{afifi2022} address the multi-illuminant WB problem as a constrained fusion problem in the sRGB domain. In particular, they pose the multi-illuminant white balance problem by obtaining a linear combination of a fixed input WB presets. In their approach, a deep neural network generates a per-pixel weighting map for the different settings. Similarly, StyleWB~\cite{kinli2023modeling} use the features of a VGG-16~\cite{vgg2015simonyan} architecture for improved performance. In this paper, we demonstrate that simply linearly combining distinct WB presets is insufficient for complex cases, as a result both MixedWB and StyleWB struggle to obtain an accurate white-balanced image.

The success of learning-based methods depends largely on the availability and quality of training and evaluation data. MixedWB and StyleWB focus on multi-illuminant sRGB WB correction but mainly rely on the RenderedWB~\cite{afifi2019color} dataset for training and Cube+~\cite{banic2017unsupervised} for testing, both of which contain only single-illuminant scenes. The Synthetic test set by Afifi et al.~\cite{afifi2022} is the only sRGB dataset for multi-illuminant scenes but lacks training samples and has a large domain gap from real images. Kim et al.~\cite{kim2021large} created the most comprehensive multi-illumination dataset by capturing RAW images of the same scene under different light sources. We repurpose this dataset to support post-WB fusion in the sRGB domain, which is our target application.

This paper introduces an efficient transformer-based architecture that fuses five WB sRGB presets non-linearly, without requiring explicit per-pixel weight maps. Additionally, we introduce a new dataset for training and evaluating multi-illuminant white balance correction in camera-rendered sRGB images. Empirically, we demonstrate that our transformer-based fusion method outperforms previous state-of-the-art approaches in WB correction by effectively blending distinct WB settings.

\section{Motivation}
MixedWB~\cite{afifi2022} and StyleWB~\cite{kinli2023modeling} hypothesize that multi-illuminant white balance can be achieved by linearly combining a set of sRGB images rendered with different preset illuminants. This assumption implies that the white-balanced pixel value should lie within the convex hull formed by the pixel values of the preset images in sRGB space. Formally, for each pixel $i$, the white balance pixel, $T_i$ is computed by:
\begin{equation}
T_i = \sum_p w_i^{p} P_i^{p},
\end{equation}
where $P$ represents the number of WB-presets values, $i$ and $p$ denote the pixel coordinates and WB preset, respectively, and $w_i$ is a per-pixel blending factor defining a convex combination of the preset values:
\begin{equation}
\sum_p w_i^{p} = 1, \quad w_i^p \geq 0, \quad \forall i, p.
\end{equation}

This assumption holds in the linear RAW space but does not necessarily apply to the non-linear sRGB domain, where both MixedWB and StyleWB operate. Figure~\ref{fig:convex_hull} illustrates this by plotting three pixels from a sample scene in the RenderedWB dataset~\cite{afifi2019color}, with WB preset values shown as dots and ground-truth values as crosses. As seen, the ground-truth pixels lie outside the convex hull formed by the presets in sRGB space, highlighting the limitations of linear blending for WB correction in this domain. To overcome this, we propose an efficient and more accurate fusion strategy using an end-to-end transformer block. Our method, represented by stars in Figure~\ref{fig:convex_hull} (g), produces values outside the convex hull, significantly closer to the ground truth. 

\section{Transformed-based WB Fusion}

Our method fuses different WB settings in a non-linear end-to-end manner. More specifically, we leverage a transformer block, as the attention mechanism inherently captures long-range spatial dependencies, making it well-suited for this task~\cite{zhang2019self, li2023flexicurve}. While vision transformers~\cite{dosovitskiy2020image} traditionally divide images into patches treated as tokens, this approach is prohibitive for high-resolution images. To improve efficiency, we adopt the channel-wise attention mechanism~\cite{zamir2022restormer} in the feature space, enabling the model to effectively aggregate information across different WB settings while maintaining computational feasibility.

First, our approach renders the RAW sensor image using five predefined WB settings --- namely {\it tungsten}, {\it fluorescent}, {\it daylight}, {\it cloudy}, and {\it shade}. These sensor responses, calibrated at the factory, can be accessed by setting the camera in manual mode and examining its metadata. The images are then rendered to sRGB using their respective WB presets. Notably, the five input images we use are readily available in most modern ISPs.

Working in sRGB with a set of pre-defined settings offers significant advantages. It transforms an extremely ill-posed problem --- obtaining the correct illuminant estimate for each pixel location --- into a more manageable one --- merging among a narrow finite set of images. Additionally, operating in sRGB rather than RAW facilitates generalization across different camera sensors whose distinct characteristics, such as sensitivities, affect RAW image formation.

Initially, we concatenate the WB presets to form a composite image, $I \in \mathbb{R}^{H \times W \times 3P}$, where $H \times W$ represents the spatial dimension of the images and $P$ denotes the number of WB presets. Given the combined WB presets, $I$, we first apply a $3 \times 3$ convolution to extract low-level features and convert the number of channels to $C$. We first use a Multi-Head Transposed Attention block~\cite{zamir2022restormer} in which we generate the {\it query} ($Q$), {\it key} ($K$), and {\it value} ($V$) projections from the convolutional features. In contrast to spatial attention, the transposed attention map $A$ is computed by swapping $Q$ and $K$ as:
\begin{equation}
    A =\text{Softmax}(K^TQ).
\end{equation}

\vspace{4pt}
After the Multi-Head Transposed Attention module, the output undergoes processing through a Feed-Forward Network~\cite{dosovitskiy2020image}, which transforms the features by operating on each location independently yet identically by formulating it as the element-wise product of two parallel paths. Finally, the output features are convolved with a $3 \times 3$ kernel to reduce the number of channels to three and produce the final sRGB white-balanced corrected image. In our experiments, we use five presets ($P{=}5$) and $C{=}15$. 

\begin{figure}
    \centering
    \begin{subfigure}{\linewidth}
        \includegraphics[width=\linewidth]{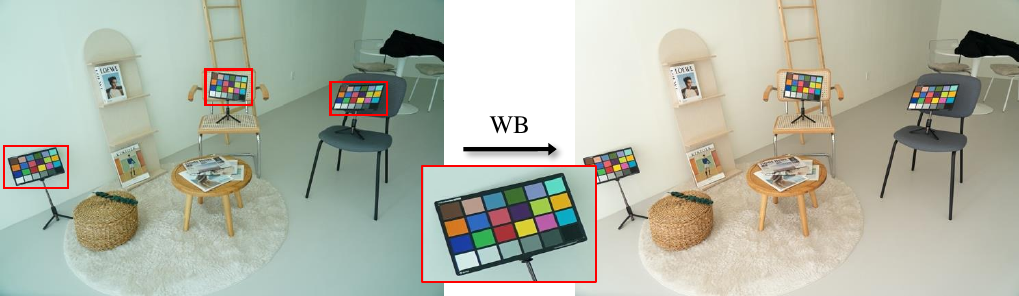}
        \caption{\textbf{Step 1}. WB correction on the single illuminant image using the Macbeth color checker and render to sRGB (GT before brightness adjustment).}
        \label{fig:dataset_creation1}
    \end{subfigure}

    \vspace{2mm}
    
    \begin{subfigure}{\linewidth}
        \includegraphics[width=\linewidth]{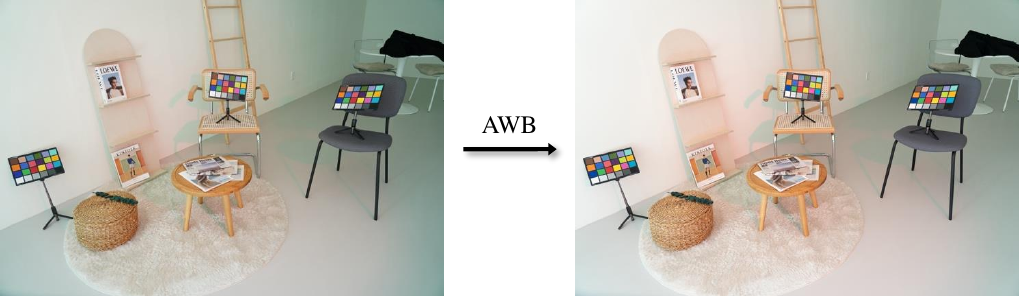}
        \caption{\textbf{Step 2}. AWB on the multiple illuminant image and render to sRGB.}
        \label{fig:dataset_creation2}

    \end{subfigure}

    \vspace{2mm}
    
    \begin{subfigure}{\linewidth}
        \includegraphics[width=\linewidth]{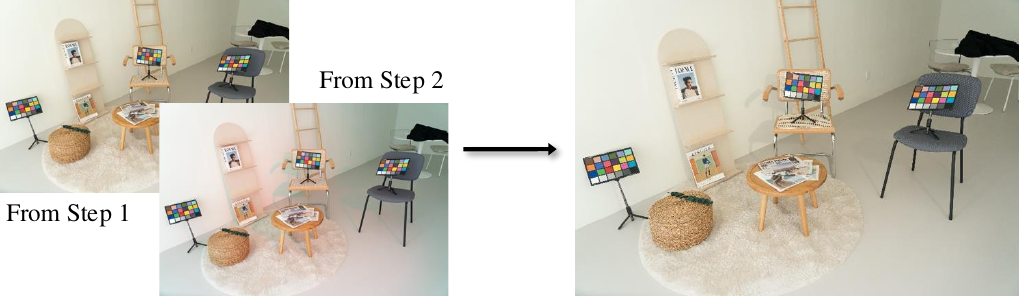}
        \caption{\textbf{Step 3}. Adjust the per-pixel brightness for the GT image.}
        \label{fig:dataset_creation3}
    \end{subfigure}
    \vspace{-5mm}
    \caption{Procedure for generating a ground truth (GT) sRGB image for scenes with multiple illuminants. We begin with the dataset from Kim et al.~\cite{kim2021large}, focused on unprocessed RAW images. All scenes are initially captured under a single illuminant, and additional illuminants are introduced individually. (a) We first compute the WB-corrected image for the single-illuminant scene using the Macbeth color checker. (b) Next, we apply AWB to the multiple-illuminant images. (c) Finally, we adjust the per-pixel brightness of the image obtained in Step 1 to generate the ground truth image. We do so by making the pixel brightness of this image match the pixel brightness from the image obtained in Step 2. }
    \label{fig:dataset_creation}
    \vspace{-5mm}
\end{figure}

\begin{figure*}[t]
\begin{center}
\includegraphics[width=0.95\textwidth]{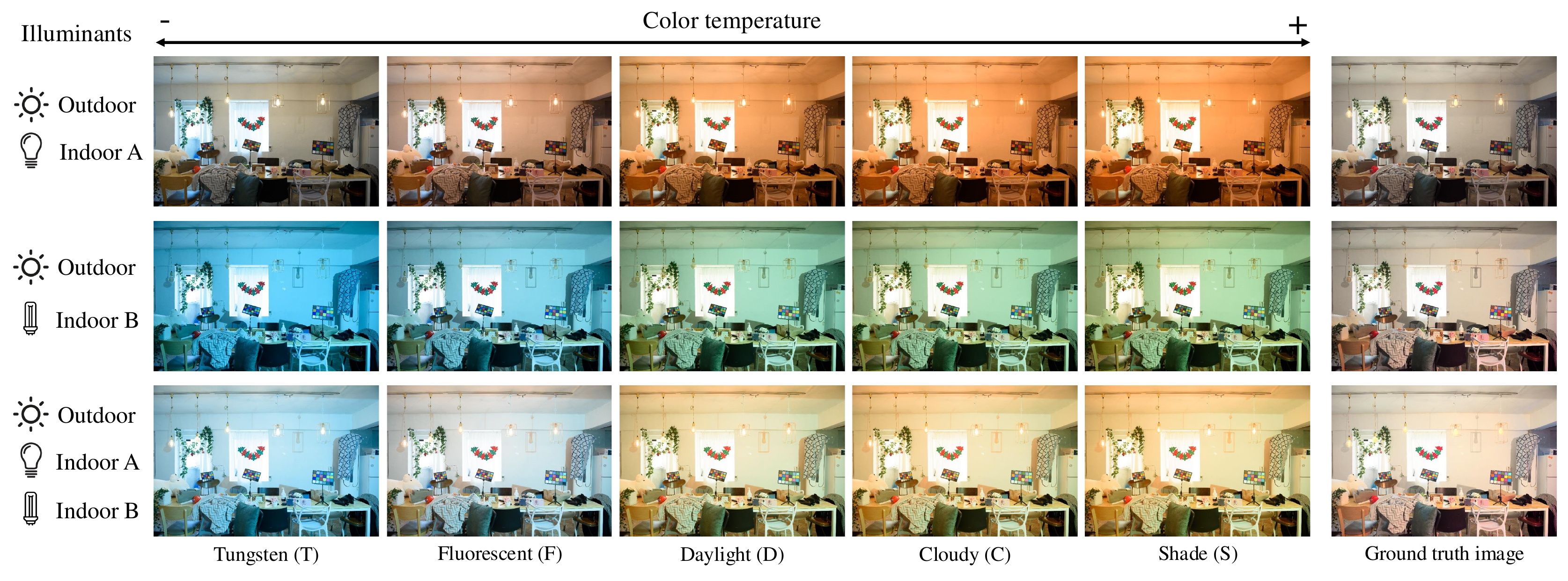}  
\end{center}
\vspace{-20pt}
\caption{Samples from our dataset, showing the same scene under varying lighting conditions and WB presets. Each column presents an image rendered with a specific WB setting alongside the ground truth. All images include outdoor lighting, while the first two rows feature two different types of indoor light, and the last row includes all three light sources (outdoor and both indoor types).}
\vspace{-12pt}
\label{fig:dataset_examples}
\end{figure*}

To summarize, our multi-illuminant WB method leverages distinct WB presets and a transformer block by fusing these presets in a nonlinear manner, allowing the model to identify regions where each setting most effectively achieves accurate white balance. Unlike MixedWB~\cite{afifi2022} and StyleWB~\cite{kinli2023modeling}, our approach does not require the inefficient per-pixel weight estimation to form the white-balanced image, handling the process in an end-to-end manner. Notably, our method has only 7.9K trainable parameters, enabling fast inference with minimal memory requirements due to the transposed attention mechanism.

\vspace{-4pt}
\section{Multi-illuminant sRGB Dataset}\label{sec:dataset}
MixedWB~\cite{afifi2022}, Style~\cite{kinli2023modeling} and, our method rely on sRGB WB presets to estimate the white-balance corrected image. However, previous state-of-the-art approaches have primarly been both trained and tested on single-illuminant datasets, such as RenderedWB~\cite{afifi2019color} and Cube+~\cite{banic2017unsupervised}, while evaluation on multi-illuminant scenes has been limited to a small synthetic dataset~\cite{afifi2022} containing only 30 scenes. To bridge this gap, we introduce a new dataset designed to support fusion-based multi-illuminant WB methods. Our dataset includes (i) five sRGB-rendered input images, each corresponding to a different WB preset, and (ii) the sRGB ground truth image with accurate WB.

Recently, Kim et al.~\cite{kim2021large} introduced the large-scale multi-illuminant (LSMI) dataset. This dataset contains RAW images of scenes with multiple cameras. For each scene, the illumination was carefully controlled. In particular, the scene was captured first under a single illumination. Afterward, additional light sources were added to the scene. This step-by-step lighting allowed a per-pixel computation in the RAW image of the contribution of each light source.  The public LSMI dataset was intended for RAW multi-illumination color constancy algorithms. This dataset serves as the basis for ours. 

In particular, we repurpose the Sony and Nikon images of the LSMI dataset \cite{kim2021large}. The total number of scenes is 955 for Nikon and 1,317 for Sony. Each scene contains either two or three illuminants. Given the DNG images of the LSMI dataset, we compute the five sRGB rendered images using Adobe Camera RAW \cite{CameraRaw} using the five standard camera WB presets: {\it daylight}, {\it cloudy}, {\it tungsten}, {\it shade}, {\it fluorescent}. We omit the Samsung Galaxy camera from the LSMI dataset due to incompatibility with Adobe Camera RAW. 
The images have resolutions of $7360 \times 4912$ and $6000 \times 4000$ for the Nikon and Sony sets, respectively. 

Figure~\ref{fig:dataset_creation} illustrates our procedure for generating ground truth white-balanced images in scenes with multiple illuminates. This step is required because the ground truth of the original LSMI dataset does not consider the differences in brightness caused by having multiple illuminant configurations for images from the same scene. To obtain the ground truth, we first apply WB correction using the color chart on the single illuminant image, as shown in step 1 of Figure~\ref{fig:dataset_creation}. However, this initial ground truth image often has less brightness than an image of the same scene with additional light sources. To correct the brightness discrepancies, we render the multi-illuminant image to sRGB using a standard AWB procedure as shown in step 2 of Figure~\ref{fig:dataset_creation}. While AWB does not produce a correct WB image under mixed lighting, it provides a reference for per-pixel brightness normalization of the single-illuminant ground truth image, as shown in step 3 of Figure~\ref{fig:dataset_creation}. This brightness adjustment assumes a Lambertian reflectance model, a fair approximation for plausible white-balanced images that maintain spatial consistency with multi-illuminant images. Our dataset provides a valuable benchmark for training and evaluating fusion-based multi-illuminant WB methods, as it introduces real-world variations absent in the synthetic test set~\cite{afifi2022}. The final dataset includes 16,284 sRGB images from the Nikon and Sony sets.

\begin{table*}[t!]
\begin{center}
\caption{\small Multi-illuminant evaluation. Quantitative evaluation on our dataset presented in Section~\ref{sec:dataset}. We compare our method against WBFlow~\cite{li2023wbflow}, SWBNet~\cite{li2023swbnet}, DeepWB~\cite{afifi2020deep}, MixedWB~\cite{afifi2022}, and StyleWB~\cite{kinli2023modeling}. All the methods are retrained on our dataset. The \dashuline{dashed line} divides the conventional sRGB WB methods and the fusion-based methods.}\label{tab:our_dataset}
\vspace{-1mm}
\footnotesize
\setlength{\tabcolsep}{6pt} 
\begin{tabular}{clccccccccc}
\toprule
& & \multicolumn{3}{c}{$\Delta$E2000} & \multicolumn{3}{c}{MSE} & \multicolumn{3}{c}{MAE} \\
\cmidrule{3-11}
Split & \multicolumn{1}{c}{Method} & Mean & Median & Trimean & Mean & Median & Trimean & Mean & Median & Trimean\\

\midrule
\multirow{6}{*}{Sony} & WBFlow~\cite{li2023wbflow} & 9.64 & 9.04 & 9.28 & 233.88 & 156.01 & 185.74 & 8.42 & 7.57 & 7.73 \\
& SWBNet~\cite{li2023swbnet} & 8.72 & 8.29 & 8.42 & 210.88 & 142.87 & 167.42 & 7.65 & 6.91 & 7.02 \\
& DeepWB~\cite{afifi2020deep} & \cellcolor{myblue}{\underline{7.53}} & \cellcolor{myblue}\underline{7.14} & \cellcolor{myblue}\underline{7.18} & \cellcolor{myblue}\underline{195.01} & \cellcolor{myblue}\underline{138.39} & \cellcolor{myblue}\underline{159.08} & \cellcolor{myblue}\underline{6.25} & \cellcolor{myblue}\underline{6.08} & \cellcolor{myblue}\underline{6.14} \\
\cdashline{2-11} \noalign{\vskip 2pt} 
& StyleWB~\cite{kinli2023modeling} & 9.46 & 8.80 & 9.00 & 230.76 & 187.40 & 193.20 & 7.89 & 7.27 & 7.54 \\
& MixedWB~\cite{afifi2022} & 8.90 & 8.21 & 8.27 & 211.88 & 153.97 & 165.03 & 6.21 & 5.74 & 5.82 \\
& Ours & \cellcolor{myred}\textbf{4.71} & \cellcolor{myred}\textbf{4.52} & \cellcolor{myred}\textbf{4.53} & \cellcolor{myred}\textbf{90.27} & \cellcolor{myred}\textbf{51.38} & \cellcolor{myred}\textbf{54.90} & \cellcolor{myred}\textbf{3.75} & \cellcolor{myred}\textbf{3.46} & \cellcolor{myred}\textbf{3.47} \\

\midrule
\multirow{6}{*}{Nikon} & WBFlow~\cite{li2023wbflow} & 9.07 & 8.57 & 8.83 & 206.18 & 181.57 & 185.03 & 6.75 & 6.25 & 6.42 \\
& SWBNet~\cite{li2023swbnet} & 8.23 & 7.83 & 7.97 & 187.61 & 164.91 & 169.30 & 6.10 & 5.69 & 5.78 \\
& DeepWB~\cite{afifi2020deep} & \cellcolor{myblue}\underline{6.66} & \cellcolor{myblue}\underline{6.40} & \cellcolor{myblue}\underline{6.47} & \cellcolor{myblue}\underline{117.60} & \cellcolor{myblue}\underline{94.14} & \cellcolor{myblue}\underline{101.70} & \cellcolor{myblue}\underline{4.66} & \cellcolor{myblue}\underline{4.23} & \cellcolor{myblue}\underline{4.44} \\
\cdashline{2-11} \noalign{\vskip 2pt} 
& StyleWB~\cite{kinli2023modeling} & 9.94 & 9.57 & 9.51 & 247.56 & 169.28 & 196.75 & 8.02 & 7.26 & 7.61 \\
& MixedWB~\cite{afifi2022} & 9.85 & 8.63 & 9.16 & 257.65 & 153.91 & 176.01 & 7.02 & 5.86 & 6.16 \\
& Ours & \cellcolor{myred}\textbf{3.84} & \cellcolor{myred}\textbf{3.55} & \cellcolor{myred}\textbf{3.58} & \cellcolor{myred}\textbf{48.91} & \cellcolor{myred}\textbf{31.59} & \cellcolor{myred}\textbf{35.88} & \cellcolor{myred}\textbf{3.13} & \cellcolor{myred}\textbf{2.83} & \cellcolor{myred}\textbf{2.82} \\

\midrule
\multirow{6}{*}{Both} & WBFlow~\cite{li2023wbflow} & 9.29 & 8.79 & 9.08 & 223.86 & 177.04 & 186.37 & 7.97 & 7.72 & 7.83 \\
& SWBNet~\cite{li2023swbnet} & 8.43 & 8.01 & 8.19 & 204.81 & 160.22 & 169.58 & 7.18 & 7.03 & 7.09 \\
& DeepWB~\cite{afifi2020deep} & \cellcolor{myblue}\underline{6.83} & \cellcolor{myblue}\underline{6.52} & \cellcolor{myblue}\underline{6.58} & \cellcolor{myblue}\underline{139.32} & \cellcolor{myblue}\underline{124.87} & \cellcolor{myblue}\underline{129.05} & \cellcolor{myblue}\underline{5.17} & \cellcolor{myblue}\underline{4.98} & \cellcolor{myblue}\underline{5.03} \\
\cdashline{2-11} \noalign{\vskip 2pt} 
& StyleWB~\cite{kinli2023modeling} & 9.67 & 8.96 & 9.13 & 234.32 & 170.38 & 190.59 & 7.92 & 7.41 & 7.59 \\
& MixedWB~\cite{afifi2022} & 9.31 & 8.45 & 8.65 & 219.51 & 148.71 & 162.40 & 6.36 & 5.60 & 5.72 \\
& Ours & \cellcolor{myred}\textbf{4.55} & \cellcolor{myred}\textbf{4.45} & \cellcolor{myred}\textbf{4.37} & \cellcolor{myred}\textbf{75.60} & \cellcolor{myred}\textbf{46.88} & \cellcolor{myred}\textbf{49.08} & \cellcolor{myred}\textbf{3.61} & \cellcolor{myred}\textbf{3.37} & \cellcolor{myred}\textbf{3.33} \\

\bottomrule
\end{tabular}
\vspace{-6mm}
\end{center}
\end{table*}

In Figure~\ref{fig:dataset_examples} we present a sample scene from our dataset and the corresponding ground truth images. Each column displays the sRGB image corresponding to a specific WB setting, arranged from left to right in order of increasing color temperature and the ground truth. The ground truth images show varying brightness levels to match the different illuminants present in each scene. All images capture the same scene but under different lighting configurations. The first row depicts the scene illuminated by both natural light from a window and an indoor light source. In the second row, the indoor light source is replaced, altering its type, position, and color temperature. The final row shows the scene with all three light sources active simultaneously.

\section{Experiments}
We evaluate the performance of our model using our dataset described in Section~\ref{sec:dataset}. The images from the Sony and Nikon cameras are organized into training (65\%), validation (15\%), and testing (20\%) splits, ensuring that all images from the same scene are in the same split. Specifically, the Sony split includes 1,092 training images, 252 validation images, and 337 testing images, while the Nikon split comprises 735 training images, 189 validation images, and 207 testing images. Furthermore, we evaluate the generalization with a cross-camera experiment and a cross-dataset evaluation on the Synthetic Multi-Illuminant test set~\cite{afifi2022}. Further, we evaluate the performance of our method on single-illuminant scenes, trained on RenderedWB~\cite{afifi2019color} and evaluated on Cube+~\cite{banic2017unsupervised}.

We evaluate the accuracy of white-balance correction using three well-established metrics: $\Delta$E2000, Mean Squared Error (MSE), and Mean Angular Error (MAE). For all metrics, we compute the mean, median, and trimean values. We highlight \colorbox{myred}{\textbf{best}} and \colorbox{myblue}{\underline{second-best}} values for each metric and separate the conventional sRGB methods of the fusion-based method with a \dashuline{dashed line}. We compare the performance of our architecture against several recent sRGB white-balance methods, including WBFlow~\cite{li2023wbflow}, SWBNet~\cite{li2023swbnet}, and DeepWB~\cite{afifi2020deep}, as well as two fusion-based approaches: MixedWB~\cite{afifi2022} and StyleWB~\cite{kinli2023modeling}. We follow the official codes and papers to train the comparison methods on our dataset. Specifically, at inference time,  we use the {\it Daylight} preset as the input image for the non-fusion-based methods, as it is the setting with the best $\Delta$E2000. For MixedWB and StyleWB we use the combination of WB presets with better performance. To fit all the models fairly we resize the Nikon images to $614 \times 920$ and the Sony images to $500 \times 700$. Note that all the color charts were masked during training to avoid learning trivial solutions.

\subsection{Implementation details}
We train our method with the Adam optimizer~\cite{kingma2014adam} with a cosine learning scheduler, starting with an initial learning rate of 1$\mathrm{e-}$3 and gradually decreasing to 1$\mathrm{e-}$5. We employ an L2 loss and select the model that achieves the lowest mean $\Delta$E2000 on the validation set.

\subsection{Results}

\paragraph{Multi-illuminant evaluation.}~Table~\ref{tab:our_dataset} summarizes mutli-illuminant performance on our dataset. The last set of rows combines Sony and Nikon images ({\it Both}). Our method substantially outperforms all other approaches across all dataset splits and evaluation metrics. Notably, due to the difficulty of the scenes, standard sRGB WB methods achieve better performance than linear fusion-based methods. Among existing works, DeepWB achieves the best performance. However, our method demonstrates a substantial improvement of 2.28 $\Delta$E2000 on the combined Sony and Nikon splits compared to DeepWB.

\begin{table*}[t!]
\begin{center}
\caption{\small Cross-camera generalization. Quantitative evaluation on cross-camera generalization of methods with unseen cameras. We present the results for all the metrics when training the models with one camera and testing it on the other split.}\label{tab:crosscamera}
\vspace{-2mm}
\footnotesize
\begin{tabular}{clccccccccc}
\toprule
& & \multicolumn{3}{c}{$\Delta$E2000} & \multicolumn{3}{c}{MSE} & \multicolumn{3}{c}{MAE} \\
\cmidrule{3-11}
Train/Test & \multicolumn{1}{c}{Method} & Mean & Median & Trimean & Mean & Median & Trimean & Mean & Median & Trimean\\
\midrule
\multirow{6}{*}{\makecell{Nikon/Sony}} 
& WBFlow~\cite{li2023wbflow} & 10.09 & 9.51 & 9.86 & 252.77 & 186.54 & 205.31 & 8.76 & 7.83 & 8.63 \\
& SWBNet~\cite{li2023swbnet} & 9.58 & 8.91 & 9.41 & 238.69 & 175.65 & 196.47 & 8.19 & 7.41 & 8.11 \\
& DeepWB~\cite{afifi2020deep} & \cellcolor{myblue}\underline{8.91} & \cellcolor{myblue}\underline{8.34} & 8.59 & \cellcolor{myblue}\underline{225.82} & \cellcolor{myblue}\underline{162.19} & \cellcolor{myblue}\underline{185.17} & 7.44 & 7.03 & 7.37 \\
\cdashline{2-11} \noalign{\vskip 2pt} 
& StyleWB~\cite{kinli2023modeling} & 9.55 & 8.92 & 8.95 & 254.30 & 217.78 & 211.05 & 7.19 & 6.52 & 6.77 \\
& MixedWB~\cite{afifi2022} & 9.17 & 8.60 & \cellcolor{myblue}\underline{8.53} & 246.42 & 207.80 & 203.91 & \cellcolor{myblue}\underline{6.89} & \cellcolor{myblue}\underline{6.27} & \cellcolor{myblue}\underline{6.45} \\
& Ours & \cellcolor{myred}\textbf{5.80} & \cellcolor{myred}\textbf{5.55} & \cellcolor{myred}\textbf{5.50} & \cellcolor{myred}\textbf{99.43} & \cellcolor{myred}\textbf{67.39} & \cellcolor{myred}\textbf{74.35} & \cellcolor{myred}\textbf{4.49} & \cellcolor{myred}\textbf{3.85} & \cellcolor{myred}\textbf{4.05}\\

\midrule

\multirow{6}{*}{\makecell{Sony/Nikon}} 
& WBFlow~\cite{li2023wbflow} & 11.02 & 10.34 & 11.00 & 272.49 & 205.75 & 225.02 & 9.81 & 8.42 & 9.60 \\
& SWBNet~\cite{li2023swbnet} & 10.54 & 9.64 & 10.15 & 266.62 & 192.16 & 213.76 & 9.06 & 8.30 & 8.70 \\
& DeepWB~\cite{afifi2020deep} & \cellcolor{myblue}\underline{9.44} & \cellcolor{myblue}\underline{8.84} & \cellcolor{myblue}\underline{9.13} & \cellcolor{myblue}\underline{239.47} & \cellcolor{myblue}\underline{175.06} & \cellcolor{myblue}\underline{197.42} & 7.99 & 7.45 & 7.86 \\
\cdashline{2-11} \noalign{\vskip 2pt}
& StyleWB~\cite{kinli2023modeling} & 10.38 & 10.08 & 10.23 & 300.06 & 246.03 & 262.13 & 8.09 & 7.63 & 7.92 \\
& MixedWB~\cite{afifi2022} & 9.96 & 9.70 & 9.76 & 289.91 & 234.54 & 254.00 & \cellcolor{myblue}\underline{7.74} & \cellcolor{myblue}\underline{7.34} & \cellcolor{myblue}\underline{7.54} \\
& Ours & \cellcolor{myred}\textbf{6.41} & \cellcolor{myred}\textbf{6.08} & \cellcolor{myred}\textbf{6.16} & \cellcolor{myred}\textbf{168.89} & \cellcolor{myred}\textbf{116.89} & \cellcolor{myred}\textbf{116.44} & \cellcolor{myred}\textbf{4.91} & \cellcolor{myred}\textbf{4.42} & \cellcolor{myred}\textbf{4.46}\\
\bottomrule
\vspace{-8mm}
\end{tabular}
\end{center}
\end{table*}

\begin{table*}[t!]
\begin{center}
\caption{\small Cross-dataset generalization. Quantitative evaluation on Synthetic test set~\cite{afifi2022}. All the methods are retrained on the combined Sony and Nikon splits of our dataset.}\label{tab:synthetic}
\vspace{-2mm}
\footnotesize
\begin{tabular}{lccccccccc}
\toprule
& \multicolumn{3}{c}{$\Delta$E2000} & \multicolumn{3}{c}{MSE} & \multicolumn{3}{c}{MAE} \\
\cmidrule{2-10}
\multicolumn{1}{c}{Method} & Mean & Median & Trimean & Mean & Median & Trimean & Mean & Median & Trimean\\
\midrule
WBFlow~\cite{li2023wbflow} & 11.86 & 10.72 & 10.82 & 728.96 & 570.52 & 634.04 & 6.08 & 6.00 & 5.69 \\
SWBNet~\cite{li2023swbnet} & 11.54 & 10.44 & 10.52 & 697.31 & 545.91 & 555.61 & 5.92 & 5.85 & 5.51 \\
DeepWB~\cite{afifi2020deep} & 10.93 & 9.82 & 10.04 & 630.60 & 486.32 & 517.09 & 5.53 & 5.55 & 5.19 \\
\cdashline{1-10} \noalign{\vskip 2pt} 
StyleWB~\cite{kinli2023modeling} & \cellcolor{myblue}\underline{9.90} & \cellcolor{myblue}\underline{9.21} & \cellcolor{myblue}\underline{9.27} & 411.88 & \cellcolor{myblue}\underline{353.97} & 365.03 & 5.21 & 4.74 & 4.82 \\
MixedWB~\cite{afifi2022} & 10.43 & 10.35 & 10.48 & \cellcolor{myblue}\underline{386.39} & 357.92 & \cellcolor{myblue}\underline{363.92} & \cellcolor{myblue}\underline{4.75} & \cellcolor{myblue}\underline{4.56} & \cellcolor{myblue}\underline{4.69} \\
Ours & \cellcolor{myred}\textbf{7.53} & \cellcolor{myred}\textbf{6.63} & \cellcolor{myred}\textbf{6.79} & \cellcolor{myred}\textbf{260.47} & \cellcolor{myred}\textbf{238.47} & \cellcolor{myred}\textbf{238.31} & \cellcolor{myred}\textbf{4.16} & \cellcolor{myred}\textbf{3.17} & \cellcolor{myred}\textbf{3.31}\\
\bottomrule
\vspace{-8mm}
\end{tabular}
\end{center}
\end{table*}

\vspace{-4mm}
\paragraph{Cross-camera generalization.}~
Our goal is to develop a WB correction method for multi-illuminant scenes that generalizes well to unseen cameras. That said, we evaluate the models trained on Sony images with Nikon images and vice versa. The results of this experiment are shown in Table~\ref{tab:crosscamera}. Our model consistently outperforms all other methods across all metrics. Notably, even when training on different splits, our approach obtains higher performance.

\vspace{-4mm}
\paragraph{Cross-dataset generalization.}~In Table~\ref{tab:synthetic}, we assess the generalization of the methods trained on our dataset on the Synthetic Multi-Illuminant test dataset~\cite{afifi2022}. Despite the domain shift between the real images and the synthetic images in this test set, our method outperforms the previous state of the art, as shown in Table~\ref{tab:synthetic}.

\begin{figure*}[t!]
\begin{center}
    \begin{subfigure}{0.16\linewidth}
        \includegraphics[width=\linewidth]{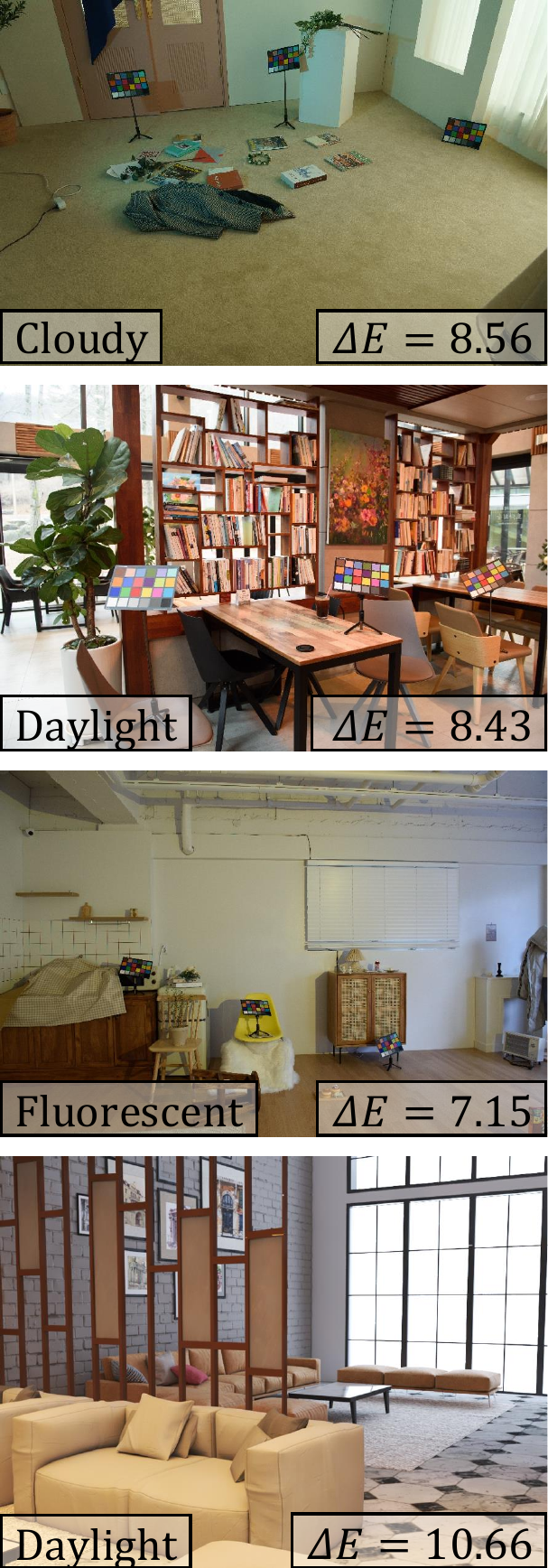}    
        \caption{Best WB setting}
    \end{subfigure}
    \begin{subfigure}{0.16\linewidth}
        \includegraphics[width=\linewidth]{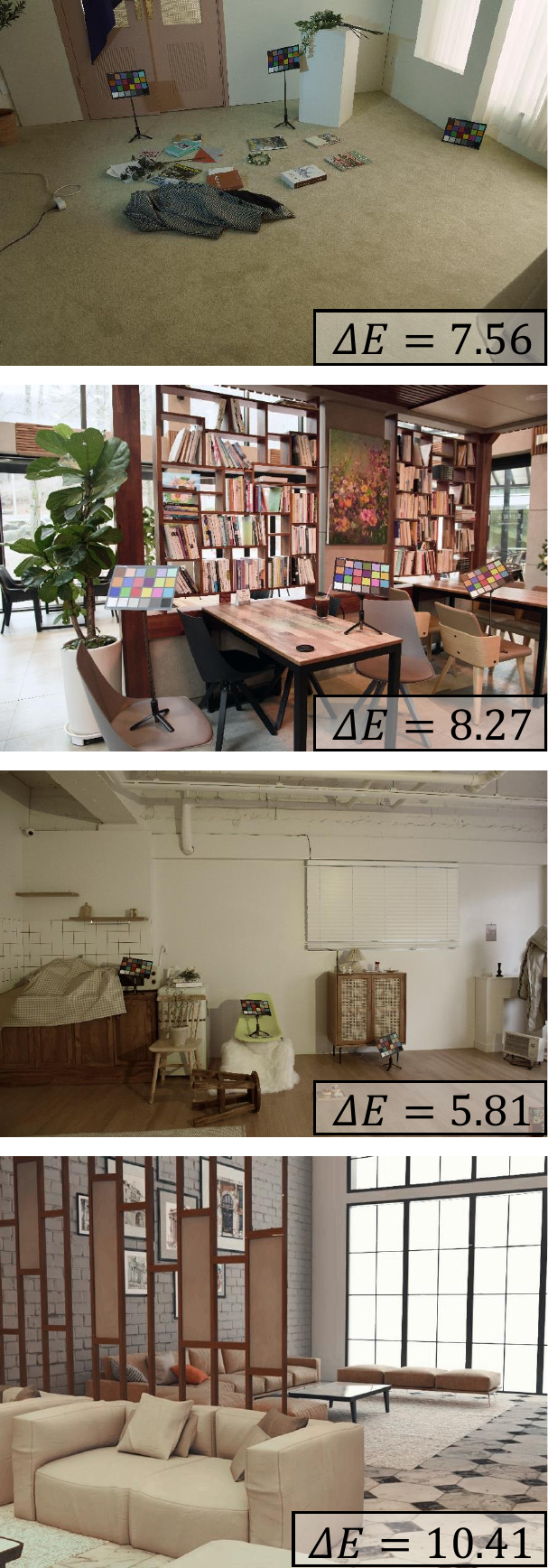}    
        \caption{DeepWB~\cite{afifi2020deep}}
    \end{subfigure}
    \begin{subfigure}{0.16\linewidth}
        \includegraphics[width=\linewidth]{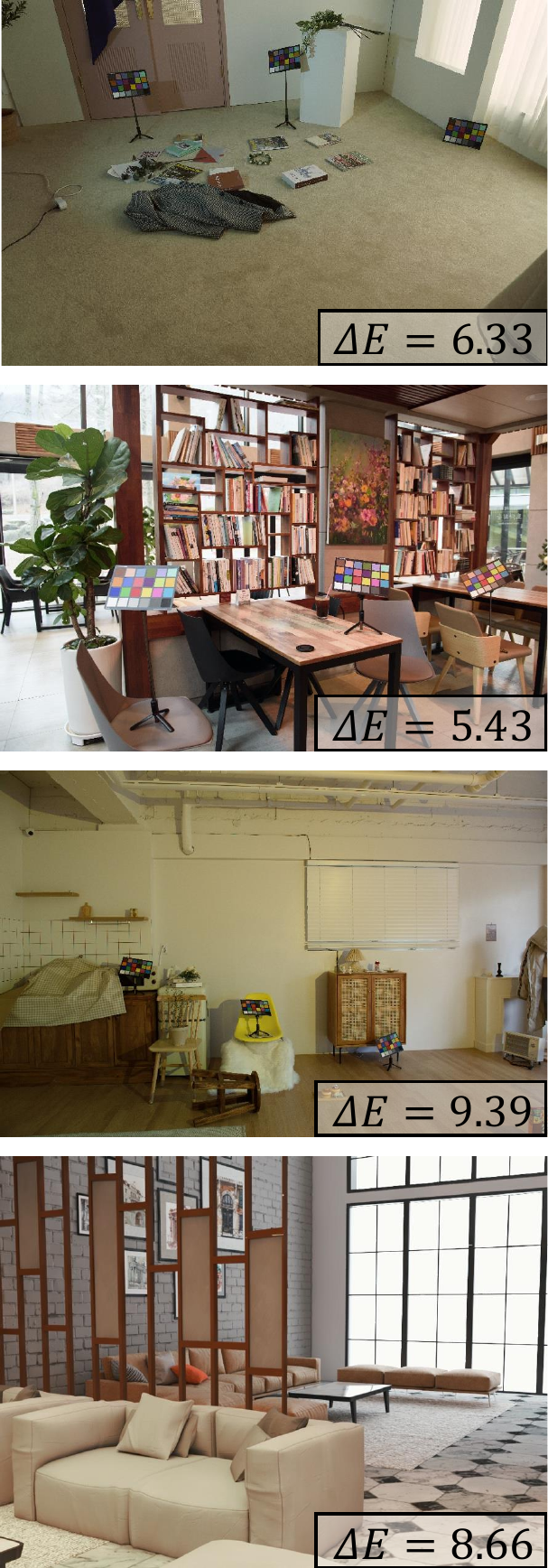}    
        \caption{MixedWB~\cite{afifi2022}}
    \end{subfigure}
    \begin{subfigure}{0.16\linewidth}
        \includegraphics[width=\linewidth]{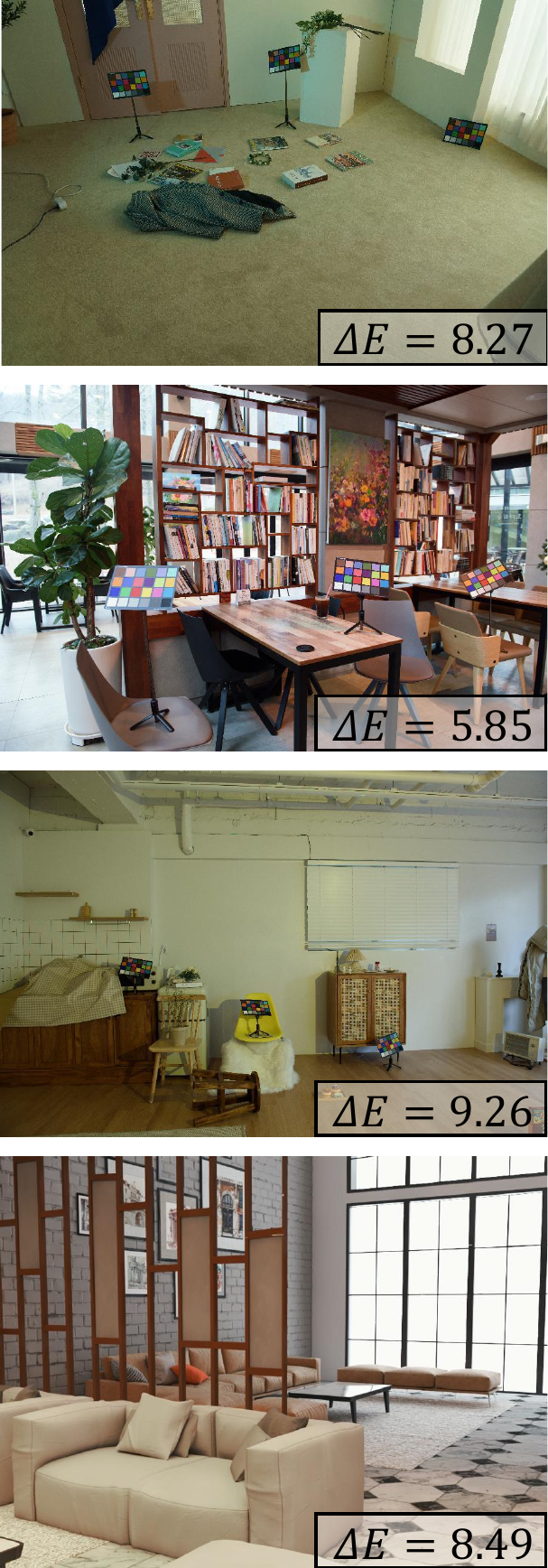}    
        \caption{StyleWB~\cite{kinli2023modeling}}
    \end{subfigure}
    \begin{subfigure}{0.16\linewidth}
        \includegraphics[width=\linewidth]{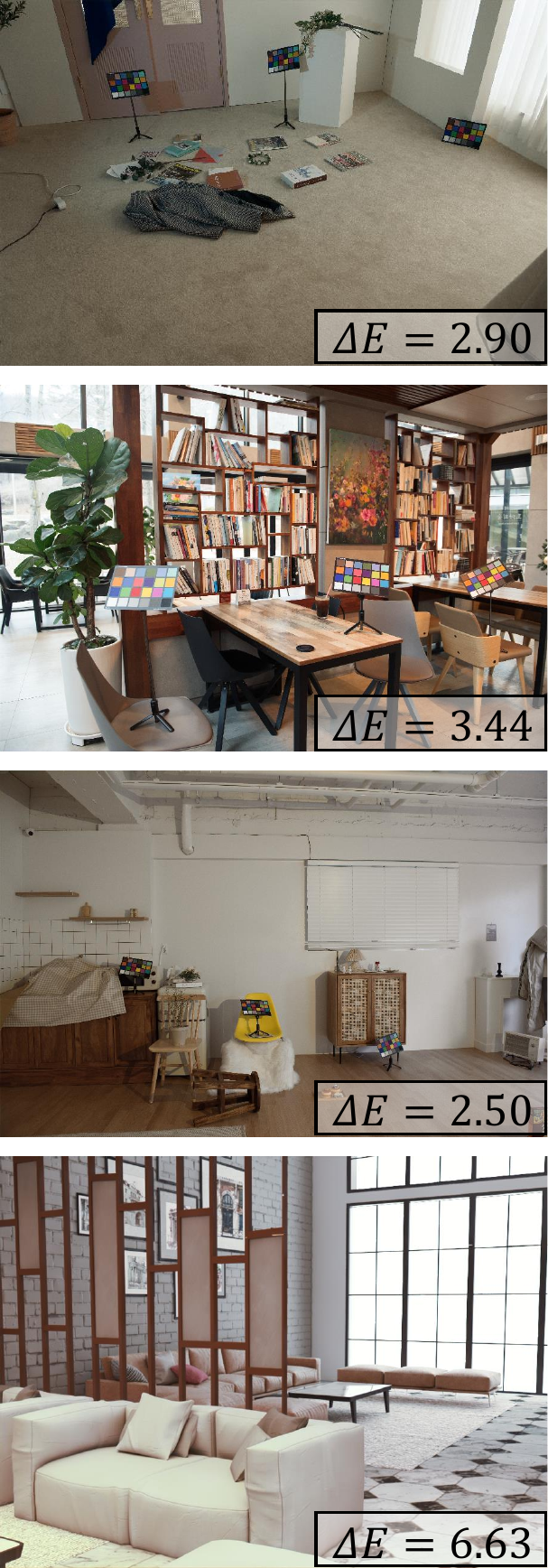}    
        \caption{Ours}
    \end{subfigure}
    \begin{subfigure}{0.16\linewidth}
        \includegraphics[width=\linewidth]{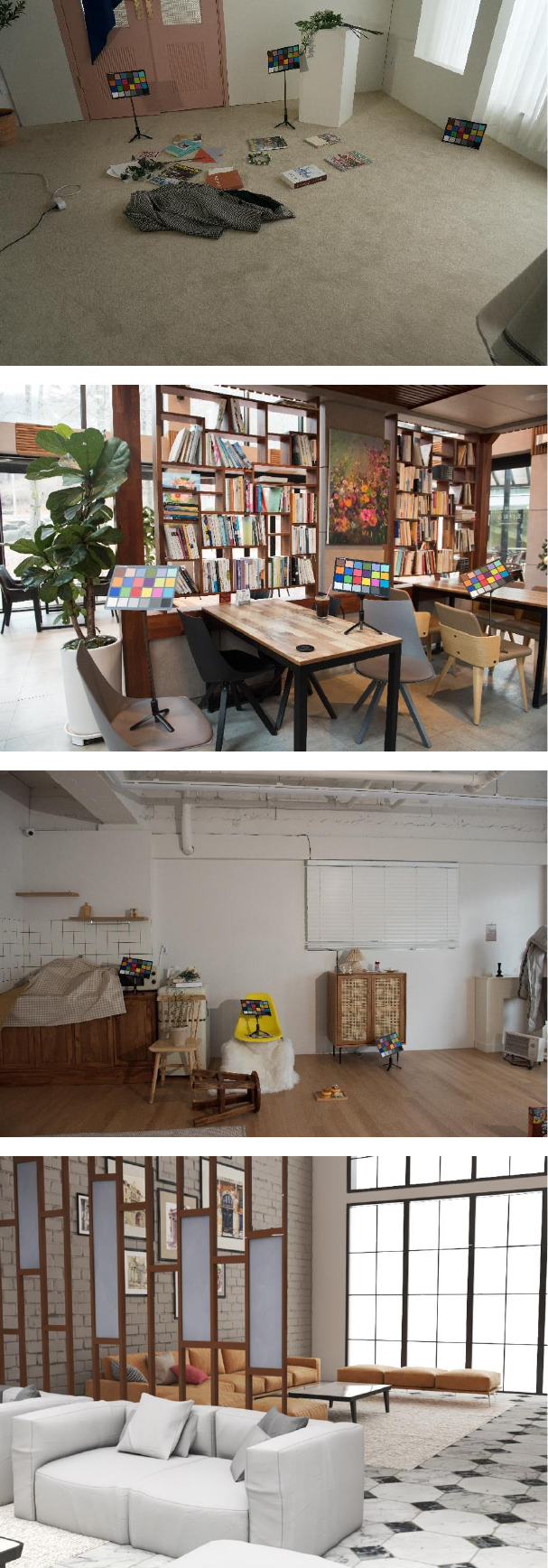}    
        \caption{Ground Truth}
    \end{subfigure}
    \vspace{-2mm}
    \caption{\small Quantitative results on our dataset and the synthetic test set~\cite{afifi2022}. Top to bottom: an image of the Sony split, two Nikon images, and the synthetic dataset. Left to right: the results of the WB preset with lowest $\Delta$E2000 (a), DeepWB~\cite{afifi2020deep} (b), MixedWB~\cite{afifi2022} (c), StyleWB~\cite{kinli2023modeling} (d), our transformer-based method (e) and the ground truth (f). In the first column (a), we show the name of the WB preset in the bottom-left corner. The $\Delta$E2000 value for each image for each image is shown in the bottom-right corner, lower values indicate higher performance.}
    \vspace{-6mm}
    \label{fig:qualitative}
\end{center}
\end{figure*}

\vspace{-4mm}
\paragraph{Single illuminant evaluation.}~Our architecture is designed for multi-illuminant white balance correction. However, we also retrain it using the same approach as MixedWB to demonstrate its superiority over the previous state of the art in image fusion for single illuminant white balance. The results on the Cube+ dataset are presented in Table~\ref{tab:cube}, with models trained on a subset of the RenderedWB dataset~\cite{afifi2019color}. In this experiment, we use the publicly available models except StyleWB which is retrained to match the training sets of the other methods. Our method consistently outperforms both standard sRGB WB methods and fusion-based approaches across all mean metrics.

\vspace{-4mm}
\paragraph{Efficiency.}~Table~\ref{tab:cube} presents the average runtime for processing 50 images with a resolution of 500$\times$700 (Nikon images) on an NVIDIA 3090 GPU. Our method has the fewest number of parameters and the fastest inference time due to the use of transposed attention. Specifically, it is nearly five times faster than DeepWB and has an order of magnitude fewer parameters compared to the other methods.

\vspace{-4mm}
\paragraph{Qualitative results.} Finally, we present qualitative results in Figure~\ref{fig:qualitative}, comparing our method to DeepWB, MixedWB, and StyleWB. The first row shows a Sony image from our dataset, the next two rows show Nikon images, and the last row presents a rendered image from the synthetic test set~\cite{afifi2022}. From left to right, we display the WB preset with the lowest $\Delta$E2000, DeepWB, MixedWB, StyleWB, our model, and the corresponding ground truth. In the first two examples, our method achieves superior results by effectively removing the color cast caused by different illuminants, visible on the floor and wooden table, respectively. In the third example, MixedWB and StyleWB perform worse than the Fluorescent WB preset, as evident on the whiteboard, illustrating that estimating a per-pixel blending map can be challenging in some cases. Lastly, the last row shows that despite the domain shift between the training data and the synthetic images from Afifi et al.~\cite{afifi2022}, our method produces results closest to the ground truth, as observed on the white sofa.

\begin{table}[t!]
\begin{center}
\caption{\small Single illuminant evaluation. Quantitative evaluation on the Cube+ dataset~\cite{banic2017unsupervised}. Results are averaged over all images.}\label{tab:cube}
\vspace{-6pt}
\footnotesize
\setlength{\tabcolsep}{4pt} 
\begin{tabular}{lccccc}
\toprule
\multicolumn{1}{c}{Method} & $\Delta$E & MSE & MAE & Param & T(ms)\\
\midrule
WBFlow~\cite{li2023wbflow} & \cellcolor{myblue}\underline{4.28} & 87.51 & \cellcolor{myblue}\underline{3.15} & 7.8M & 927\\
SWBNet~\cite{li2023swbnet} & \cellcolor{myblue}\underline{4.28} & \cellcolor{myblue}\underline{74.35} & \cellcolor{myblue}\underline{3.15} & 2.1M & 693\\
DeepWB~\cite{afifi2020deep} & 4.59 & 80.46 & 3.45 & 4.4M & 459\\
\cdashline{1-6} \noalign{\vskip 2pt} 
StyleWB*~\cite{kinli2023modeling} & 4.88 & 95.82 & 3.61 & 15.5M & 1893 \\
MixedWB~\cite{afifi2022} & 5.03 & 168.38 & 4.20 & 1.3M & 980\\
Ours & \cellcolor{myred}\textbf{4.19} & \cellcolor{myred}\textbf{68.33} & \cellcolor{myred}\textbf{2.98} & 7.9K & 179 \\
\bottomrule
\vspace{-20pt}
\end{tabular}
\end{center}
\end{table}

\begin{table}[t!]
\begin{center}
\caption{\small Study on the number of presets. Models are trained and evaluated on our dataset's combined Sony and Nikon splits.}\label{tab:ablation}
\vspace{-6pt}
\footnotesize
\setlength{\tabcolsep}{4pt} 
\begin{tabular}{lcccccc}
\toprule
& \multicolumn{3}{c}{$\Delta$E2000} & \multicolumn{3}{c}{MAE} \\
\cmidrule{2-7}
\multicolumn{1}{c}{Method} & Mean & Median & Trimean & Mean & Median & Trimean \\
\midrule
Daylight & 5.47 & 5.23 & 5.19 & 4.27 & 3.88 & 4.02 \\
DST & 5.18 & 4.94 & 4.96 & 4.06 & 3.74 & 3.76 \\
All presets & 4.55 & 4.45 & 4.37 & 3.61 & 3.37 & 3.33 \\
\bottomrule
\vspace{-30pt}
\end{tabular}
\end{center}
\end{table}

\subsection{Ablation Study}
\paragraph{Standard sRGB WB vs fusion-based.}~As seen in Tables~\ref{tab:our_dataset} and~\ref{tab:crosscamera}, in complex cases such as those in our dataset, linear fusion-based methods struggle to produce accurate WB images and are outperformed by regular sRGB WB methods. To further evaluate the effectiveness of our non-linear fusion strategy, we conduct additional experiments to compare it against regular sRGB WB. Specifically, we modify our transformer block to use only the {\it Daylight} preset as input and, separately, to use three input presets (DST) by adjusting the first convolutional layer. Table~\ref{tab:ablation} reports the $\Delta$E2000 and MAE values for these two cases, as well as for our proposed approach using all five presets in the Sony and Nikon splits of our dataset ({\it Both} in Table~\ref{tab:our_dataset}). Even when using only the {\it Daylight} preset, our model outperforms previous state-of-the-art methods. However, we observe that performance improves as more presets are incorporated. This suggests that the model benefits from additional presets, as they provide more information on how different illuminants interact under varying color temperatures.


\section{Conclusion}
In this work, we introduced a transformer-based approach for WB correction in multi-illuminant scenes by efficiently fusing WB presets. Our method outperforms previous linear fusion-based approaches and achieves robust generalization in unseen cameras and datasets without the need to compute per-pixel blending maps. By leveraging transposed attention, our approach demonstrated superior adaptability, and efficiency compared to the state of the art. Additionally, we introduced an sRGB dataset tailored for multi-illuminant WB correction. We anticipate that our dataset will support further advancement in multi-illuminant white-balance research.

\section*{Acknowledgements}
DSL, LH, and JVC were supported by Grant PID2021-128178OB-I00 funded by MCIN/AEI/10.13039/ 501100011033 and by ERDF "A way of making Europe", by the Departament de Recerca i Universitats from Generalitat de Catalunya with reference 2021SGR01499, and by the Generalitat de Catalunya CERCA Program. DSL also acknowledges the FPI grant from the Spanish Ministry of Science and Innovation (PRE2022-101525). LH was also supported by the Ramon y Cajal grant RYC2019-027020-I. This work was also partially supported by the grant Càtedra ENIA UAB-Cruïlla (TSI-100929-2023-2) from the Ministry of Economic Affairs and Digital Transition of Spain. KGD and MSB were supported by the CFREF (VISTA) program, an NSERC Discovery Grant. The Canada Research Chair program also supported MSB.

{
    \small
    \bibliographystyle{ieeenat_fullname}
    \bibliography{main}

\begin{thebibliography}{55}
\providecommand{\natexlab}[1]{#1}
\providecommand{\url}[1]{\texttt{#1}}
\expandafter\ifx\csname urlstyle\endcsname\relax
  \providecommand{\doi}[1]{doi: #1}\else
  \providecommand{\doi}{doi: \begingroup \urlstyle{rm}\Url}\fi

\bibitem[{Adobe Systems}()]{CameraRaw}
{Adobe Systems}.
\newblock Adobe {C}amera {R}aw.
\newblock Computer software.

\bibitem[Afifi and Brown(2020{\natexlab{a}})]{afifi2020deep}
Mahmoud Afifi and Michael~S Brown.
\newblock Deep white-balance editing.
\newblock In \emph{CVPR}, 2020{\natexlab{a}}.

\bibitem[Afifi and Brown(2020{\natexlab{b}})]{afifi2020interactive}
Mahmoud Afifi and Michael~S Brown.
\newblock Interactive white balancing for camera-rendered images.
\newblock In \emph{CIC}, 2020{\natexlab{b}}.

\bibitem[Afifi et~al.(2019{\natexlab{a}})Afifi, Price, Cohen, and Brown]{afifi2019color}
Mahmoud Afifi, Brian Price, Scott Cohen, and Michael~S Brown.
\newblock When color constancy goes wrong: Correcting improperly white-balanced images.
\newblock In \emph{CVPR}, 2019{\natexlab{a}}.

\bibitem[Afifi et~al.(2019{\natexlab{b}})Afifi, Punnappurath, Abdelhamed, Karaimer, Abuolaim, and Brown]{afifi2019tuning}
Mahmoud Afifi, Abhijith Punnappurath, Abdelrahman Abdelhamed, Hakki~Can Karaimer, Abdullah Abuolaim, and Michael~S Brown.
\newblock Color temperature tuning: Allowing accurate post-capture white-balance editing.
\newblock In \emph{CIC}, 2019{\natexlab{b}}.

\bibitem[Afifi et~al.(2022)Afifi, Brubaker, and Brown]{afifi2022}
Mahmoud Afifi, Marcus~A Brubaker, and Michael~S Brown.
\newblock Auto white-balance correction for mixed-illuminant scenes.
\newblock In \emph{WACV}, 2022.

\bibitem[Bani{\'c} et~al.(2017)Bani{\'c}, Ko{\v{s}}{\v{c}}evi{\'c}, and Lon{\v{c}}ari{\'c}]{banic2017unsupervised}
Nikola Bani{\'c}, Karlo Ko{\v{s}}{\v{c}}evi{\'c}, and Sven Lon{\v{c}}ari{\'c}.
\newblock Unsupervised learning for color constancy.
\newblock \emph{arXiv preprint arXiv:1712.00436}, 2017.

\bibitem[Barron and Tsai(2017)]{DBLP:conf/cvpr/BarronT17}
Jonathan~T. Barron and Yun{-}Ta Tsai.
\newblock Fast {F}ourier color constancy.
\newblock In \emph{CVPR}, 2017.

\bibitem[Beigpour et~al.(2013)Beigpour, Riess, Van De~Weijer, and Angelopoulou]{beigpour2013multi}
Shida Beigpour, Christian Riess, Joost Van De~Weijer, and Elli Angelopoulou.
\newblock Multi-illuminant estimation with conditional random fields.
\newblock \emph{TIP}, 23\penalty0 (1):\penalty0 83--96, 2013.

\bibitem[Bianco and Cusano(2019)]{bianco2019quasi}
Simone Bianco and Claudio Cusano.
\newblock Quasi-unsupervised color constancy.
\newblock In \emph{CVPR}, 2019.

\bibitem[Brainard and Freeman(1997)]{brainard1997bayesian}
David~H Brainard and William~T Freeman.
\newblock Bayesian color constancy.
\newblock \emph{JOSA A}, 1997.

\bibitem[Buchsbaum(1980)]{BUCHSBAUM19801}
George Buchsbaum.
\newblock A spatial processor model for object color perception.
\newblock \emph{J. Frank. Inst.}, 1980.

\bibitem[Chakrabarti et~al.(2008)Chakrabarti, Hirakawa, and Zickler]{chakrabarti2008color}
Ayan Chakrabarti, Keigo Hirakawa, and Todd Zickler.
\newblock Color constancy beyond bags of pixels.
\newblock In \emph{CVPR}, 2008.

\bibitem[Cheng et~al.(2014)Cheng, Prasad, and Brown]{cheng2014illuminant}
Dongliang Cheng, Dilip~K Prasad, and Michael~S Brown.
\newblock Illuminant estimation for color constancy: {W}hy spatial-domain methods work and the role of the color distribution.
\newblock \emph{JOSA A}, 2014.

\bibitem[Dosovitskiy et~al.(2021)Dosovitskiy, Beyer, Kolesnikov, Weissenborn, Zhai, Unterthiner, Dehghani, Minderer, Heigold, Gelly, Uszkoreit, and Houlsby]{dosovitskiy2020image}
Alexey Dosovitskiy, Lucas Beyer, Alexander Kolesnikov, Dirk Weissenborn, Xiaohua Zhai, Thomas Unterthiner, Mostafa Dehghani, Matthias Minderer, Georg Heigold, Sylvain Gelly, Jakob Uszkoreit, and Neil Houlsby.
\newblock An image is worth 16x16 words: Transformers for image recognition at scale.
\newblock \emph{ICLR}, 2021.

\bibitem[Finlayson(2013)]{finlayson2013corrected}
Graham~D Finlayson.
\newblock Corrected-moment illuminant estimation.
\newblock In \emph{ICCV}, 2013.

\bibitem[Finlayson and Trezzi(2004)]{finlayson2004shades}
Graham~D Finlayson and Elisabetta Trezzi.
\newblock Shades of gray and colour constancy.
\newblock In \emph{CIC}, 2004.

\bibitem[Finlayson et~al.(1995)Finlayson, Funt, and Barnard]{finlayson1995color}
Graham~D Finlayson, Brian~V Funt, and Kobus Barnard.
\newblock Color constancy under varying illumination.
\newblock In \emph{ICCV}, 1995.

\bibitem[Finlayson et~al.(1997)Finlayson, Hubel, and Hordley]{finlayson1997color}
Graham~D Finlayson, Paul~M Hubel, and Steven Hordley.
\newblock Color by correlation.
\newblock In \emph{CIC}, 1997.

\bibitem[Gehler et~al.(2008)Gehler, Rother, Blake, Minka, and Sharp]{gehler2008bayesian}
Peter~Vincent Gehler, Carsten Rother, Andrew Blake, Tom Minka, and Toby Sharp.
\newblock Bayesian color constancy revisited.
\newblock In \emph{CVPR}, 2008.

\bibitem[Gijsenij et~al.(2011{\natexlab{a}})Gijsenij, Gevers, and Van De~Weijer]{DBLP:journals/tip/GijsenijGW11}
Arjan Gijsenij, Theo Gevers, and Joost Van De~Weijer.
\newblock Computational color constancy: Survey and experiments.
\newblock \emph{TIP}, 20\penalty0 (9):\penalty0 2475--2489, 2011{\natexlab{a}}.

\bibitem[Gijsenij et~al.(2011{\natexlab{b}})Gijsenij, Gevers, and Van De~Weijer]{gijsenij2011improving}
Arjan Gijsenij, Theo Gevers, and Joost Van De~Weijer.
\newblock Improving color constancy by photometric edge weighting.
\newblock \emph{IEEE TPAMI}, 34\penalty0 (5):\penalty0 918--929, 2011{\natexlab{b}}.

\bibitem[Gijsenij et~al.(2012)Gijsenij, Lu, and Gevers]{gijsenij2011color}
Arjan Gijsenij, Rui Lu, and Theo Gevers.
\newblock Color constancy for multiple light sources.
\newblock \emph{TIP}, 2012.

\bibitem[Hernandez-Juarez et~al.(2020)Hernandez-Juarez, Parisot, Busam, Leonardis, Slabaugh, and McDonagh]{hernandez2020multi}
Daniel Hernandez-Juarez, Sarah Parisot, Benjamin Busam, Ales Leonardis, Gregory Slabaugh, and Steven McDonagh.
\newblock A multi-hypothesis approach to color constancy.
\newblock In \emph{CVPR}, 2020.

\bibitem[Hsu et~al.(2008)Hsu, Mertens, Paris, Avidan, and Durand]{hsu2008light}
Eugene Hsu, Tom Mertens, Sylvain Paris, Shai Avidan, and Fr{\'e}do Durand.
\newblock Light mixture estimation for spatially varying white balance.
\newblock In \emph{ACM SIGGRAPH}, 2008.

\bibitem[Hu et~al.(2017)Hu, Wang, and Lin]{DBLP:conf/cvpr/HuWL17}
Yuanming Hu, Baoyuan Wang, and Stephen Lin.
\newblock {FC}4: Fully convolutional color constancy with confidence-weighted pooling.
\newblock In \emph{CVPR}, 2017.

\bibitem[Hussain and Akbari(2018)]{hussain2018color}
Md~Akmol Hussain and Akbar~Sheikh Akbari.
\newblock Color constancy algorithm for mixed-illuminant scene images.
\newblock \emph{IEEE Access}, 2018.

\bibitem[Joze and Drew(2013)]{joze2013exemplar}
Hamid Reza~Vaezi Joze and Mark~S Drew.
\newblock Exemplar-based color constancy and multiple illumination.
\newblock \emph{PAMI}, 36\penalty0 (5):\penalty0 860--873, 2013.

\bibitem[Kim et~al.(2021)Kim, Kim, Nam, Lee, Lee, Kang, Lee, Yoo, Han, and Kim]{kim2021large}
Dongyoung Kim, Jinwoo Kim, Seonghyeon Nam, Dongwoo Lee, Yeonkyung Lee, Nahyup Kang, Hyong-Euk Lee, ByungIn Yoo, Jae-Joon Han, and Seon~Joo Kim.
\newblock Large scale multi-illuminant ({LSMI}) dataset for developing white balance algorithm under mixed illumination.
\newblock In \emph{ICCV}, 2021.

\bibitem[Kim et~al.(2024)Kim, Kim, Yu, and Kim]{kim2024attentive}
Dongyoung Kim, Jinwoo Kim, Junsang Yu, and Seon~Joo Kim.
\newblock Attentive illumination decomposition model for multi-illuminant white balancing.
\newblock In \emph{CVPR}, 2024.

\bibitem[Kingma and Ba(2014)]{kingma2014adam}
Diederik~P Kingma and Jimmy Ba.
\newblock Adam: A method for stochastic optimization.
\newblock \emph{arXiv preprint arXiv:1412.6980}, 2014.

\bibitem[K{\i}nl{\i} et~al.(2023)K{\i}nl{\i}, Y{\i}lmaz, {\"O}zcan, and K{\i}ra{\c{c}}]{kinli2023modeling}
Furkan K{\i}nl{\i}, Do{\u{g}}a Y{\i}lmaz, Bar{\i}{\c{s}} {\"O}zcan, and Furkan K{\i}ra{\c{c}}.
\newblock Modeling the lighting in scenes as style for auto white-balance correction.
\newblock In \emph{WACV}, 2023.

\bibitem[Land and McCann(1971)]{land1971lightness}
Edwin~H Land and John~J McCann.
\newblock Lightness and retinex theory.
\newblock \emph{JOSA A}, 1971.

\bibitem[Li et~al.(2023{\natexlab{a}})Li, Guo, Zhou, Ai, Feng, and Loy]{li2023flexicurve}
Chongyi Li, Chunle Guo, Shangchen Zhou, Qiming Ai, Ruicheng Feng, and Chen~Change Loy.
\newblock Flexicurve: Flexible piecewise curves estimation for photo retouching.
\newblock In \emph{CVPR}, 2023{\natexlab{a}}.

\bibitem[Li et~al.(2023{\natexlab{b}})Li, Kang, and Ming]{li2023wbflow}
Chunxiao Li, Xuejing Kang, and Anlong Ming.
\newblock Wbflow: Few-shot white balance for srgb images via reversible neural flows.
\newblock In \emph{International Joint Conference on Artificial Intelligence}, 2023{\natexlab{b}}.

\bibitem[Li et~al.(2023{\natexlab{c}})Li, Kang, Zhang, and Ming]{li2023swbnet}
Chunxiao Li, Xuejing Kang, Zhifeng Zhang, and Anlong Ming.
\newblock Swbnet: A stable white balance network for s{RGB} images.
\newblock In \emph{AAAI}, 2023{\natexlab{c}}.

\bibitem[Lo et~al.(2021)Lo, Chang, Chiu, Huang, Chen, Chang, and Jou]{lo2021clcc}
Yi-Chen Lo, Chia-Che Chang, Hsuan-Chao Chiu, Yu-Hao Huang, Chia-Ping Chen, Yu-Lin Chang, and Kevin Jou.
\newblock {CLCC}: Contrastive learning for color constancy.
\newblock In \emph{CVPR}, 2021.

\bibitem[Lou et~al.(2015)Lou, Gevers, Hu, and Lucassen]{lou2015color}
Zhongyu Lou, Theo Gevers, Ninghang Hu, and Marcel~P Lucassen.
\newblock Color constancy by deep learning.
\newblock In \emph{BMVC}, 2015.

\bibitem[Oh and Kim(2017)]{oh2017approaching}
Seoung~Wug Oh and Seon~Joo Kim.
\newblock Approaching the computational color constancy as a classification problem through deep learning.
\newblock \emph{PR}, 61:\penalty0 405--416, 2017.

\bibitem[Qian et~al.(2019)Qian, Kamarainen, Nikkanen, and Matas]{qian2019finding}
Yanlin Qian, Joni-Kristian Kamarainen, Jarno Nikkanen, and Jiri Matas.
\newblock On finding gray pixels.
\newblock In \emph{CVPR}, 2019.

\bibitem[Sapiro(1999)]{sapiro1999color}
Guillermo Sapiro.
\newblock Color and illuminant voting.
\newblock \emph{PAMI}, 1999.

\bibitem[Shi et~al.(2016)Shi, Loy, and Tang]{shi2016deep}
Wu Shi, Chen~Change Loy, and Xiaoou Tang.
\newblock Deep specialized network for illuminant estimation.
\newblock In \emph{ECCV}, 2016.

\bibitem[Sidorov(2019)]{sidorov2019conditional}
Oleksii Sidorov.
\newblock Conditional {GANs} for multi-illuminant color constancy: Revolution or yet another approach?
\newblock In \emph{CVPRW}, 2019.

\bibitem[Simonyan and Zisserman(2015)]{vgg2015simonyan}
Karen Simonyan and Andrew Zisserman.
\newblock Very deep convolutional networks for large-scale image recognition.
\newblock In \emph{ICLR}, 2015.

\bibitem[Van De~Weijer et~al.(2007)Van De~Weijer, Gevers, and Gijsenij]{van2007edge}
Joost Van De~Weijer, Theo Gevers, and Arjan Gijsenij.
\newblock Edge-based color constancy.
\newblock \emph{TIP}, 16\penalty0 (9):\penalty0 2207--2214, 2007.

\bibitem[Vaswani et~al.(2017)Vaswani, Shazeer, Parmar, Uszkoreit, Jones, Gomez, Kaiser, and Polosukhin]{vaswani2017attention}
Ashish Vaswani, Noam Shazeer, Niki Parmar, Jakob Uszkoreit, Llion Jones, Aidan~N Gomez, {\L}ukasz Kaiser, and Illia Polosukhin.
\newblock Attention is all you need.
\newblock In \emph{NeurIPS}, 2017.

\bibitem[Vazquez et~al.(2009)Vazquez, P{\'a}rraga, Vanrell, and Baldrich]{vazquez2009color}
Javier Vazquez, C~Alejandro P{\'a}rraga, Maria Vanrell, and Ramon Baldrich.
\newblock Color constancy algorithms: Psychophysical evaluation on a new dataset.
\newblock \emph{Journal of Imaging Science and Technology}, 1\penalty0 (3):\penalty0 1, 2009.

\bibitem[Vazquez-Corral et~al.(2011)Vazquez-Corral, Vanrell, Baldrich, and Tous]{vazquez2011color}
Javier Vazquez-Corral, Maria Vanrell, Ramon Baldrich, and Francesc Tous.
\newblock Color constancy by category correlation.
\newblock \emph{TIP}, 21\penalty0 (4):\penalty0 1997--2007, 2011.

\bibitem[Vr{\v{s}}nak et~al.(2022{\natexlab{a}})Vr{\v{s}}nak, Domislovi{\'c}, Suba{\v{s}}i{\'c}, and Lon{\v{c}}ari{\'c}]{vrvsnak2022autoencoder}
Donik Vr{\v{s}}nak, Ilija Domislovi{\'c}, Marko Suba{\v{s}}i{\'c}, and Sven Lon{\v{c}}ari{\'c}.
\newblock Autoencoder-based training for multi-illuminant color constancy.
\newblock \emph{JOSA A}, 2022{\natexlab{a}}.

\bibitem[Vr{\v{s}}nak et~al.(2022{\natexlab{b}})Vr{\v{s}}nak, Domislovi{\'c}, Suba{\v{s}}i{\'c}, and Lon{\v{c}}ari{\'c}]{vrvsnak2022illuminant}
Donik Vr{\v{s}}nak, Ilija Domislovi{\'c}, Marko Suba{\v{s}}i{\'c}, and Sven Lon{\v{c}}ari{\'c}.
\newblock Illuminant segmentation for multi-illuminant scenes using latent illumination encoding.
\newblock \emph{Signal Process. Image Commun.}, 2022{\natexlab{b}}.

\bibitem[Xu et~al.(2020)Xu, Liu, Hou, Liu, and Qiu]{xu2020end}
Bolei Xu, Jingxin Liu, Xianxu Hou, Bozhi Liu, and Guoping Qiu.
\newblock End-to-end illuminant estimation based on deep metric learning.
\newblock In \emph{CVPR}, 2020.

\bibitem[Yue and Wei(2024)]{yue2024robust}
Shuwei Yue and Minchen Wei.
\newblock Robust pixel-wise illuminant estimation algorithm for images with a low bit-depth.
\newblock \emph{Optics Express}, 32\penalty0 (15):\penalty0 26708--26718, 2024.

\bibitem[Zamir et~al.(2022)Zamir, Arora, Khan, Hayat, Khan, and Yang]{zamir2022restormer}
Syed~Waqas Zamir, Aditya Arora, Salman Khan, Munawar Hayat, Fahad~Shahbaz Khan, and Ming-Hsuan Yang.
\newblock Restormer: Efficient transformer for high-resolution image restoration.
\newblock In \emph{CVPR}, 2022.

\bibitem[Zhang et~al.(2019)Zhang, Goodfellow, Metaxas, and Odena]{zhang2019self}
Han Zhang, Ian Goodfellow, Dimitris Metaxas, and Augustus Odena.
\newblock Self-attention generative adversarial networks.
\newblock In \emph{ICML}, 2019.

\bibitem[Zini et~al.(2022)Zini, Buzzelli, Bianco, and Schettini]{zini2022cocoa}
Simone Zini, Marco Buzzelli, Simone Bianco, and Raimondo Schettini.
\newblock Cocoa: combining color constancy algorithms for images and videos.
\newblock \emph{IEEE Transactions on Computational Imaging}, 8:\penalty0 795--807, 2022.

\end{thebibliography}
}

\end{document}